\documentclass[11pt]{article}

\usepackage[svgnames, x11names, table]{xcolor} 

\usepackage{acl}
\usepackage{times}
\usepackage{latexsym}
\usepackage[T1]{fontenc}
\usepackage[utf8]{inputenc}
\usepackage{microtype}
\usepackage{inconsolata}
\usepackage{float}      
\usepackage{graphicx}
\usepackage{multirow}
\usepackage{booktabs}
\usepackage{tabularx}
\usepackage{makecell}
\usepackage{fontawesome5}
\usepackage{hyperref}
\usepackage{pifont}
\usepackage{makecell}
\usepackage{listings}   
\usepackage{hyperref}
\usepackage{rotating} 
\usepackage{pdflscape}
\usepackage{afterpage} 
\usepackage{xltabular}
\usepackage{booktabs}
\usepackage{caption}
\usepackage{amsmath}
\usepackage{ragged2e}
\usepackage{stfloats}
\usepackage{pifont}
\usepackage{xspace}
\usepackage[utf8]{inputenc} 
\DeclareUnicodeCharacter{00A0}{ } 
\DeclareUnicodeCharacter{3000}{ } 
\usepackage{titletoc}
\usepackage{tabularx}
\usepackage{threeparttable}
\usepackage{pifont} 
\usepackage[table]{xcolor}
\newcommand{\xmark}{\ding{55}} 
\newcolumntype{C}{>{\centering\arraybackslash}X}
\usepackage{tcolorbox}
\tcbuselibrary{listings, skins, breakable} 
\usepackage{fancyhdr}
\usepackage{graphicx}
\usepackage{xcolor}
\usepackage{graphicx}
\fancypagestyle{firstpage}{
  \fancyhf{} 
  \fancyhead[L]{
    \raisebox{0.3\height}{%
      \includegraphics[width=60pt,height=19pt]{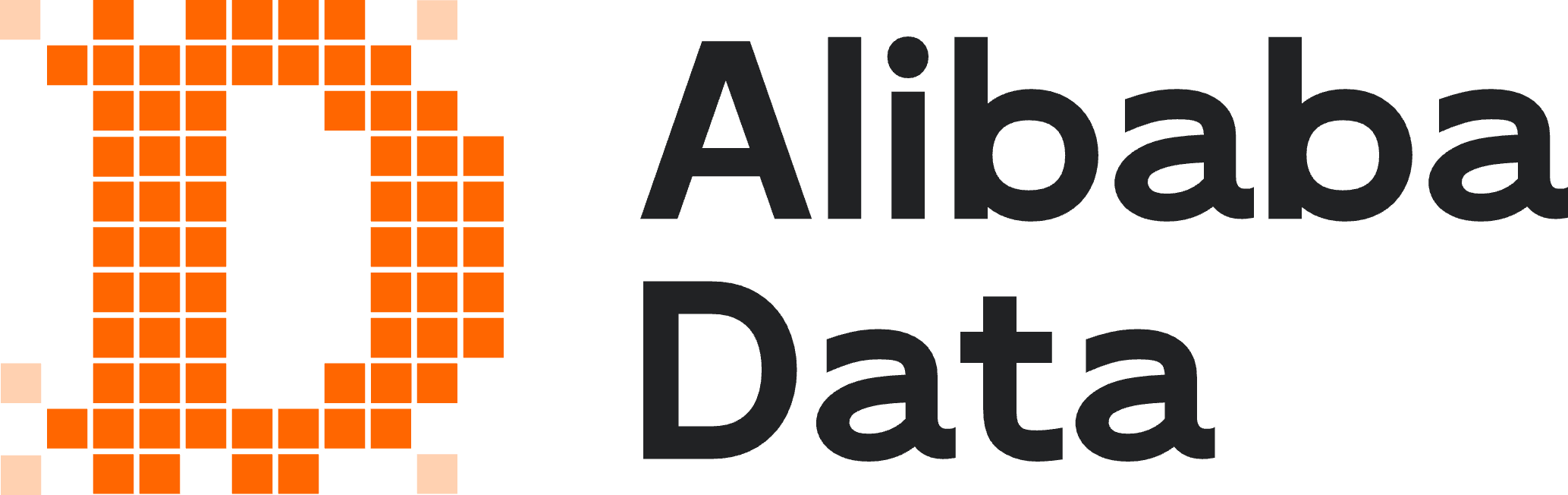}%
      \hspace{1em}%
      \includegraphics[height=21pt]{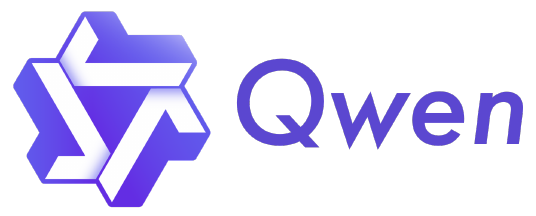}%
    }%
  }
  \fancyhead[R]{
    \raisebox{0.3\height}{%
      \includegraphics[height=21pt]{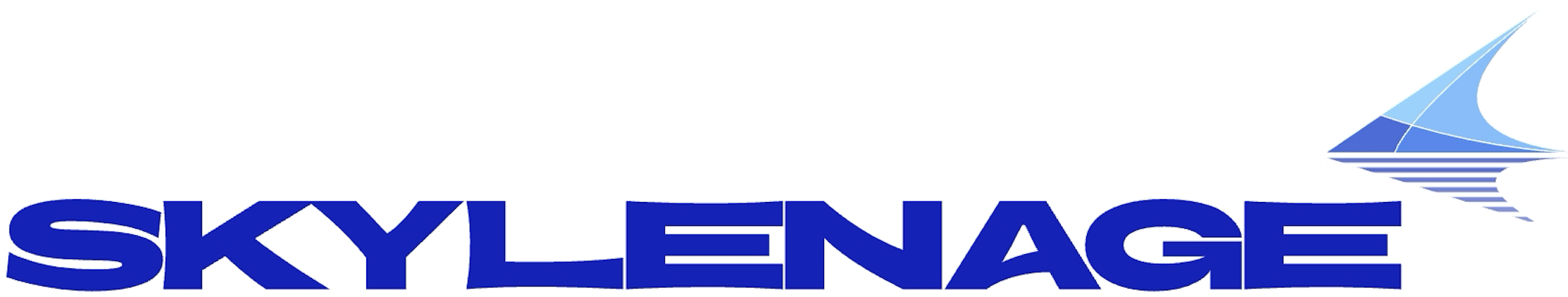}%
    }%
  }
}

\newtcolorbox{PromptBox}[1][]{
    title={\textbf{\small User Input}},
    colback=WhiteSmoke,
    colframe=RoyalBlue3,
    boxrule=1.5pt,
    arc=1mm,
    enhanced,
    breakable,   
    fonttitle=\bfseries,
    #1
}

\definecolor{yescolor}{RGB}{0, 150, 0}
\definecolor{nocolor}{RGB}{200, 200, 200}
\newcommand{\yes}{\textcolor{yescolor}{\ding{51}}}

\definecolor{summarybg}{gray}{0.95}
\definecolor{oursbg}{RGB}{235, 245, 255} 

\newcommand{\PLawBench}{\textsc{PLawBench}\xspace}

\title{\PLawBench: A Rubric-Based Benchmark for Evaluating LLMs in Real-World Legal Practice}

\author{
 \textbf{Yuzhen Shi\textsuperscript{1}\thanks{These authors contributed equally to this work.}},
 \textbf{Huanghai Liu\textsuperscript{2,5}\footnotemark[1]},
 \textbf{Yiran Hu\textsuperscript{6}\footnotemark[1]},
 \textbf{Gaojie Song\textsuperscript{3}},
 \textbf{Xinran Xu\textsuperscript{3,4}},
 \textbf{Yubo Ma\textsuperscript{2}},
 \textbf{Tianyi Tang\textsuperscript{2}},
 \textbf{Li Zhang \textsuperscript{7}},
 \\
 \textbf{Qingjing Chen\textsuperscript{8}},
 \textbf{Di Feng\textsuperscript{3}},
 \textbf{Wenbo Lv\textsuperscript{3}},
 \textbf{Weiheng Wu\textsuperscript{3}},
 \textbf{Kexin Yang\textsuperscript{2}},
 \textbf{Sen Yang\textsuperscript{2}},
 \textbf{Wei Wang\textsuperscript{1}},
 \textbf{Rongyao Shi\textsuperscript{3}},
 \\
 \textbf{Yuanyang Qiu\textsuperscript{3}},
 \textbf{Yuemeng Qi\textsuperscript{3}},
 \textbf{Jingwen Zhang\textsuperscript{3}},
 \textbf{Xiaoyu Sui\textsuperscript{3}},
 \textbf{Yifan Chen\textsuperscript{1}},
 \textbf{Yi Zhang\textsuperscript{3}},
 \textbf{An Yang\textsuperscript{2}},
 \textbf{Bowen Yu\textsuperscript{2}},
 \\
 \textbf{Dayiheng Liu\textsuperscript{2}},
 \textbf{Junyang Lin\textsuperscript{2}},
 \textbf{Weixing Shen\textsuperscript{5}},
 \textbf{Bing Zhao\textsuperscript{1}\thanks{Corresponding authors.}},
 \textbf{Charles L.A. Clarke\textsuperscript{6}},
 \textbf{Hu Wei\textsuperscript{1}\footnotemark[2]}
\\
 \textsuperscript{1}Alibaba Group
 \textsuperscript{2}Qwen Team%
\raisebox{-0.2\height}{%
  \includegraphics[height=1.2em]{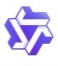}%
}, Alibaba Group
\textsuperscript{3}Skylenage
\\
 \textsuperscript{4}Shanghai Jiao Tong University
  \textsuperscript{5}Tsinghua University
  \\
  \textsuperscript{6}University of Waterloo
 \textsuperscript{7}University of Pittsburgh
  \textsuperscript{8}University of Bologna
\\
}

\begin{document}
\maketitle
\thispagestyle{firstpage}

\vspace{2em}
\noindent

\begin{abstract}

As large language models (LLMs) are increasingly applied to legal domain-specific tasks, evaluating their ability to perform legal work in real-world settings has become essential. However, existing legal benchmarks rely on simplified and highly standardized tasks, failing to capture the ambiguity, complexity, and reasoning demands of real legal practice. Moreover, prior evaluations often adopt coarse, single-dimensional metrics and do not explicitly assess fine-grained legal reasoning. To address these limitations, we introduce \textbf{\PLawBench}, a \textbf{P}ractical \textbf{Law} \textbf{Bench}mark designed to evaluate LLMs in realistic legal practice scenarios. Grounded in real-world legal workflows, \PLawBench models the core processes of legal practitioners through three task categories: public legal consultation, practical case analysis, and legal document generation. These tasks assess a model's ability to identify legal issues and key facts, perform structured legal reasoning, and generate legally coherent documents. \PLawBench comprises 850 questions across 13 practical legal scenarios, with each question accompanied by expert-designed evaluation rubrics, resulting in approximately 12,500 rubric items for fine-grained assessment. Using an LLM-based evaluator aligned with human expert judgments, we evaluate 10 state-of-the-art LLMs. Experimental results show that none achieves strong performance on \PLawBench, revealing substantial limitations in the fine-grained legal reasoning capabilities of current LLMs and highlighting important directions for future evaluation and development of legal LLMs. Data is available at: \url{https://github.com/skylenage/PLawbench}.

\end{abstract}

\section{Introduction}

With the rapid advancement of AI technologies, large language models (LLMs) have been increasingly applied to a wide range of domain-specific applications, such as legal~\cite{chen2024survey, padiu2024extent}, medical~\cite{zhou2023survey}, and finance~\cite{li2023large}. In these knowledge-intensive fields, real-world problems often require the integration of extensive prior knowledge together with complex logical reasoning. Enabling LLMs to handle professional tasks in such domains, therefore, necessitates the coordinated use of multiple capabilities, including knowledge retrieval~\cite{pipitone2024legalbench}, reasoning~\cite{yao2025elevating}, planning~\cite{li2025legalagentbench}, and decision making~\cite{chen2024alignment}.

In the legal domain, LLMs have been explored as tools to assist with real-world legal tasks traditionally performed by legal practitioners. Practical legal scenarios are highly diverse and complex, requiring the ability to retrieve and apply legal knowledge, identify and clarify legally relevant facts, plan multi-step problem-solving strategies, conduct structured legal reasoning~\cite{shi2025legalreasoner}, and verify and analyze conclusions~\cite{aba1992legal, doyle2025if}. This raises a fundamental question: \textbf{To what extent can LLMs handle complex legal tasks and reason like human practitioners?}

\begin{figure*}[!h]
    \includegraphics[width=\textwidth]{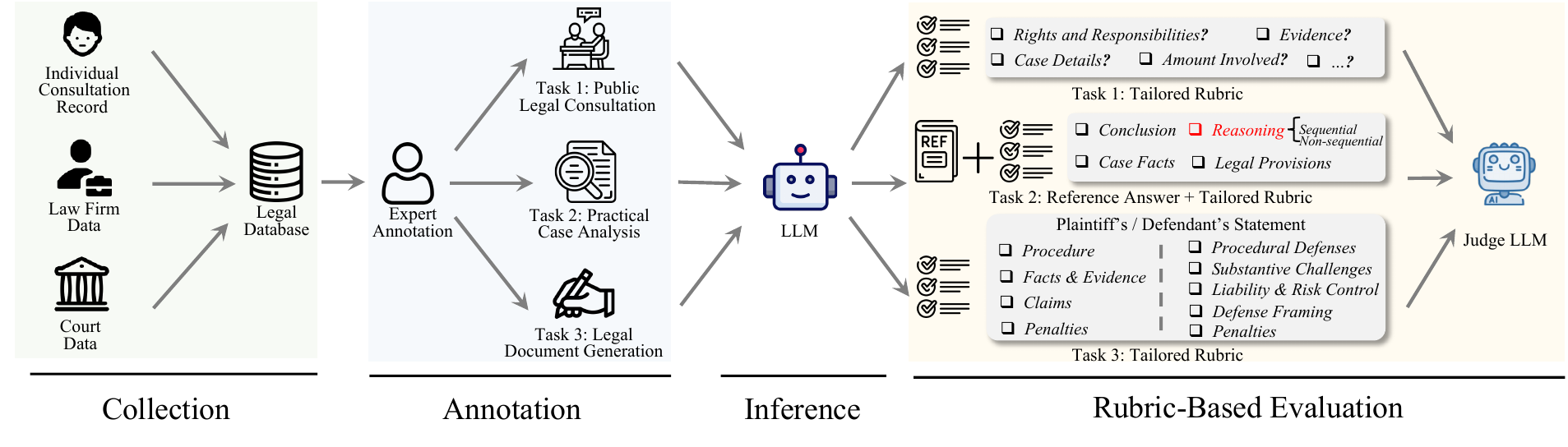}
    \caption{Overall framework of \PLawBench illustrates a four-step pipeline: collecting multi-source legal data, expert annotation into three task types, LLM-based inference on these tasks, and rubric-based evaluation of the LLM outputs by a judge model.}
    \label{fig：main}
\end{figure*}

Prior research has attempted to evaluate LLMs on legal tasks~\cite{cui2023survey, hu2025j, liu2025jurex, fei-etal-2024-lawbench, li2025lexrag}.
However, the construction of these benchmarks often fails to reflect the practical demands of realistic legal workflows.
Specifically, existing work suffers from the following three limitations:
\paragraph{\ding{172} Lack of Realistic Task Construction.}
    
    Some previous benchmarks are directly derived from judicial or bar examination questions
    (e.g., LAIW~\cite{dai-etal-2025-laiw}, LEXAM~\cite{fan2025lexam}),
    while others contain tasks that are deliberately simplified
    (e.g., LegalAgentBench~\cite{li2025legalagentbench}, Legalbench~\cite{guha2023legalbench}). In real-world legal settings, user queries are frequently ambiguous, incomplete, or strategically framed,
    with key facts omitted or obscured.
    Only by incorporating such characteristics can evaluations more accurately capture
    LLM performance when acting as legal practitioners.

\paragraph{\ding{173} Insufficient Modeling of Legal Reasoning.}
    
    Logical and structured legal reasoning steps are essential when LLMs handle legal problems.
    Prior benchmarks claim to assess reasoning abilities. LexEval~\cite{li2024lexeval} evaluates ``Logic Inference'' capability while their designs overlook the distinctive nature of legal reasoning.
    Simple syllogistic reasoning~\cite{deng2023syllogistic} or IRAC
    (Issue, Rule, Application, and Conclusion)-style~\cite{jang2025pilot}
    evaluations primarily capture surface structure rather than substantive legal inference~\cite{zhang2025thinking}.
    Even recent benchmarks, such as PRBENCH~\cite{akyurek2025prbench}, rely on general-purpose scoring frameworks
    that are not tailored to the unique reasoning mechanisms of the legal domain.
    As a result, these approaches fail to adequately assess the complex,
    context-dependent reasoning required in real legal practice.

\paragraph{\ding{174} Coarse and Uniform Evaluation Metrics.}
    
    Moreover, existing evaluations typically rely on coarse criteria such as whether the final conclusion is correct,
    whether relevant statutes are cited, or whether a syllogistic structure is followed
    ~\cite{fei-etal-2024-lawbench, li2024lexeval, wald2018contextual}.
    However, such uniform evaluation metrics fail to capture the diversity and complexity of legal practice
    across different scenarios.
    Fine-grained and case-specific evaluation rubrics are essential for realistically assessing
    legal practitioner LLMs.

Due to these limitations, existing benchmarks have not achieved a comprehensive, in-depth,
or a realistic assessment of LLM capabilities in legal practice.
To address this gap, we propose a \textbf{P}ractical \textbf{Law} \textbf{Bench}mark
(\textbf{\PLawBench}) designed to evaluate LLMs acting as legal practitioners under
realistic, knowledge-intensive, and reasoning-driven conditions.
This work has three main contributions:

\paragraph{\ding{182} Authentic Simulation of Legal Practice.} We evaluate LLMs using a hierarchical framework adapted from authentic legal workflows, comprising three levels: \textit{Public Legal Consultation}, \textit{Practical Case Analysis}, and \textit{Legal Document Generation} (see Figure~\ref{fig：main})~\footnote{Adapted from Maughan and Webb \cite{maughan2005lawyering}: \textit{Consultation} corresponds to "Interviewing" (Ch.5), \textit{Case Analysis} to "Case Management" (Ch.10), and \textit{Document Generation} to "Drafting" (Ch.7-8).}. To mirror the cognitive challenges of real-world practice, we adapt tasks from actual cases and deliberately incorporate realistic noise — such as vague queries, emotional narratives, and omitted facts. This design moves beyond idealized benchmarks to rigorously test whether models can distill relevant legal facts from the ambiguity inherent in professional environments.
\paragraph{\ding{183} Fine-Grained Reasoning Steps.}
Beyond evaluating the final conclusions, our benchmark explicitly incorporates fine-grained legal reasoning steps into task design and evaluation. This allows us to examine whether LLMs can perform multi-stage legal reasoning, including issue identification, fact clarification, legal analysis, and conclusion validation, rather than relying on shallow pattern matching or surface-level reasoning.

\paragraph{\ding{184} Task-Specific Rubric.}
Our framework employs personalized, task-specific rubrics to assess substantive legal reasoning beyond mere outcomes. Legal experts follow a two-stage annotation process: first defining a reasoning-oriented framework for each task type, then tailoring specific criteria to individual case scenarios. This approach ensures evaluation is both principled and context-sensitive, enabling a fine-grained and realistic assessment of LLM performance in practical settings.

\section{Related Work}

Existing research has proposed a variety of benchmarks to evaluate AI systems in legal domains, primarily focusing on models' ability to understand and apply legal knowledge. Early efforts assess legal AI systems through tasks such as legal information extraction and legal reasoning. Representative benchmarks include JEC-QA~\cite{zhong2020jecqa}, which evaluates multi-hop reasoning based on Chinese bar examination questions; CaseHOLD~\cite{zheng2024casehold}, which focuses on identifying legal holdings in U.S. appellate decisions;~\cite{zhang2025llms}, which evaluates the identification of overruling relationships in U.S. Supreme Court cases; and SARA~\cite{blair-stanek2023sara}, which examines statutory reasoning in U.S. tax law. Many of these benchmarks adopt syllogistic reasoning structures or legal writing frameworks such as IRAC.

\begin{figure*}[b] 
    \centering
    \makebox[\textwidth][c]{
        \includegraphics[width=1.2\linewidth]{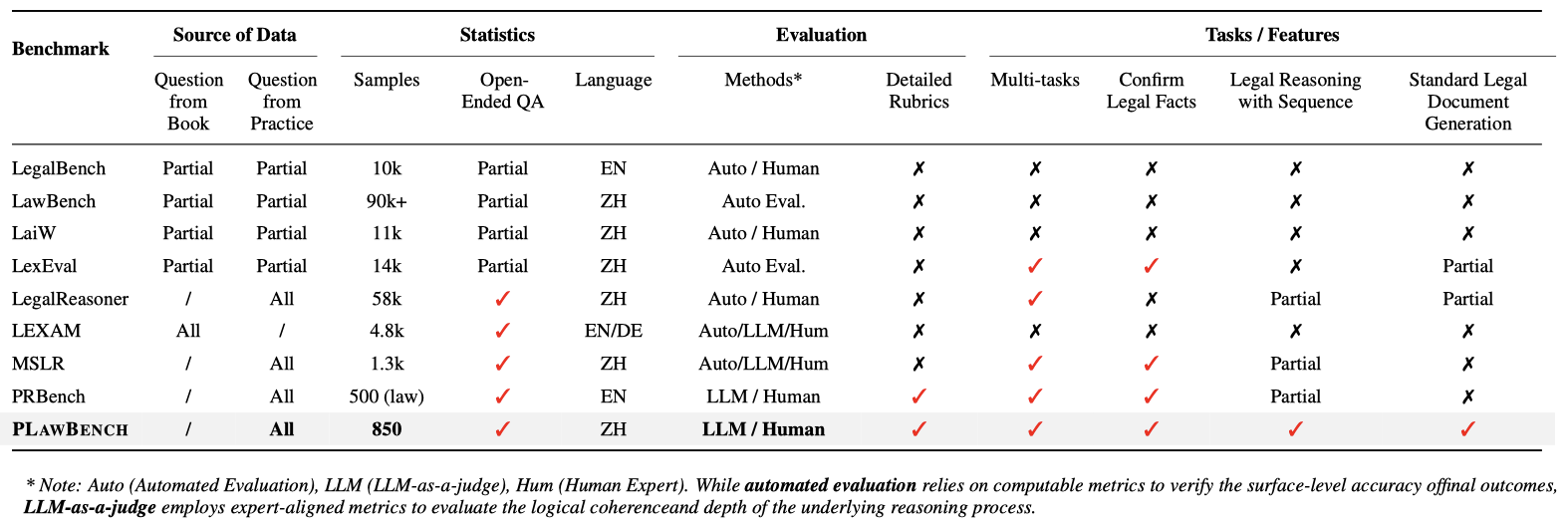}
    }
    \caption{Comprehensive Comparison of \PLawBench with Existing Legal Benchmarks.}
    \label{fig:compare}
\end{figure*}

More recently, comprehensive benchmarks have been introduced to provide multi-dimensional evaluations of legal AI systems. For example, LawBench~\cite{fei-etal-2024-lawbench} evaluates legal LLMs across cognitive levels, including memorization, understanding, and application; LegalBench~\cite{guha2023legalbench} comprises a wide range of tasks spanning multiple forms of legal reasoning; LexEval~\cite{li2024lexeval}, the largest Chinese legal benchmark to date, covers diverse legal tasks at scale; and LAiW~\cite{dai-etal-2025-laiw} explicitly structures legal reasoning into layered components aligned with legal practice logic.

Despite their contributions, existing benchmarks often fail to reflect the complexities of real-world legal practice. First, tasks like statutory memorization or exam-based QA (e.g., LawBench~\cite{fei-etal-2024-lawbench}) diverge significantly from authentic scenarios where practitioners must distill relevant facts from ambiguous, incomplete client narratives. Second, current benchmarks typically isolate interdependent tasks — such as separating statutory retrieval from judgment prediction — ignoring the holistic reasoning chain required in practice. Furthermore, evaluation metrics for document generation (e.g., ROUGE-L) fail to capture professional legal standards. In contrast, our benchmark integrates three representative authentic tasks with fine-grained, expert-designed rubrics, ensuring comprehensive and practically relevant evaluation. A comprehensive comparison of \PLawBench with existing legal benchmarks is illustrated in Figure \ref{fig:compare}, while a more in-depth analysis of existing benchmarks is provided in Appendix \ref{app:survey}.

\begin{table*}[t]
\centering
\small
\resizebox{0.8\linewidth}{!}{%
\begin{tabular}{p{4.5cm} p{6.3cm}}
\toprule
\textbf{Dimension} & \textbf{What It Evaluates} \\
\midrule
Issue \& Fact Identification & Identification of legal issues and material facts \\
Legal Reasoning & Logical inference connecting facts and legal rules \\
Legal Knowledge Application & Context-aware application of legal provisions \\
Procedural \& Strategic Awareness & Procedural compliance and litigation strategy \\
Claim \& Outcome Construction & Formulation of claims, defenses, and remedies \\
Professional Norms \& Compliance & Adherence to legal and ethical standards \\
\bottomrule
\end{tabular}
}
\caption{Rubric dimensions defined based on practical legal tasks.}
\label{tab:unified_rubric}
\end{table*}

\section{\PLawBench}
\subsection{Framework}
\paragraph{Overall Framework} As illustrated in Figure 1, we organize legal tasks into three hierarchical levels based on the real-world workflow of legal practitioners: public legal consultation, practical case analysis, and legal document generation. These three tasks are designed to evaluate distinct but complementary capabilities of LLM-based agents: identifying legal issues and clarifying key facts, analyzing disputes through legal reasoning, and summarizing and presenting legal claims in a structured and formalized manner, respectively. 



\paragraph{Rubric Construction}

Our rubric design adopts a risk-tiered perspective. It is grounded in principles emphasized by the EU AI Act's~\cite{act2024eu} risk classification framework, AI auditing standards, and legal risk management practices: the risks posed by AI systems must be evaluated in relation to specific tasks, concrete contexts, and affected stakeholders, rather than relying solely on abstract accuracy metrics~\cite{hagan2024measuring, calloway2025multidimensional}.

In legal settings, traditional notions of ``accuracy'' are insufficient to capture task complexity. In many scenarios, information may be factually correct yet operationally unusable, legally accurate but high-risk in terms of rights implications, or correct but lacking essential procedural guidance — all of which can result in substantial harm to users.

Accordingly, we abstract task-level rubric items into six unified evaluation dimensions, each corresponding to a fundamental component of legal practice: (1) Issue and Fact Identification, (2) Legal Reasoning, (3) Legal Knowledge Application, (4) Procedural and Strategic Awareness, (5) Claim and Outcome Construction, and (6) Professional Norms and Compliance as shown in Table \ref{tab:unified_rubric}. These dimensions are designed to capture both substantive legal reasoning and practical decision-making, rather than surface-level task performance. Building on the six unified evaluation dimensions, we map them onto specific legal task settings and integrate them into rubric design to ensure that risk assessment aligns with task‑specific legal risk points.

\begin{figure*}[t]
    \centering
    \includegraphics[width=\textwidth]{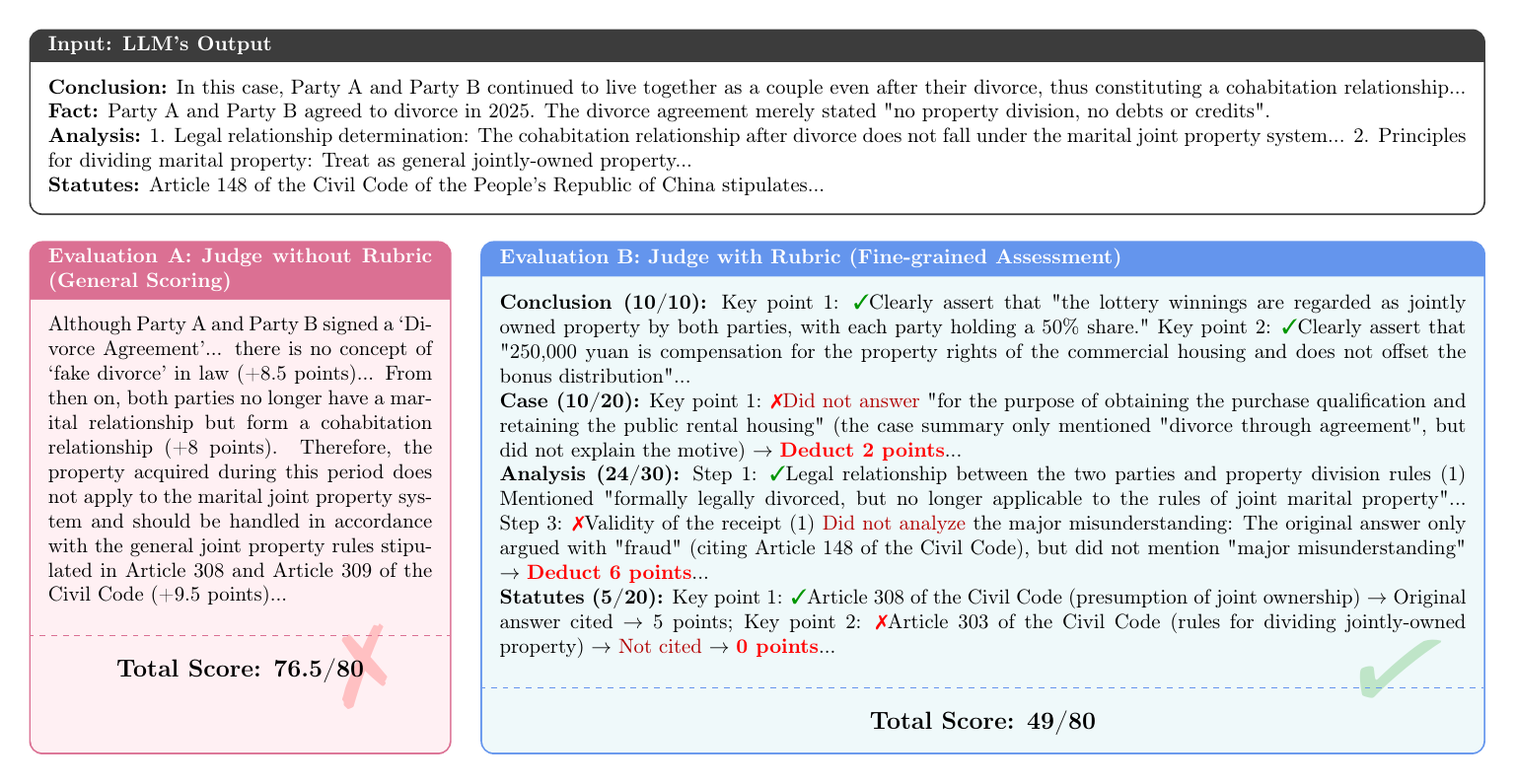}
    \caption{A contrasting example: rubric-based approach (\textbf{Evaluation B}) can identify situations that appear accurate on the surface but are actually flawed in the reasoning process, while rubric-free approach (\textbf{Evaluation A}) cannot do this, potentially exposing users to significant legal risks in practice.}
    \label{fig:example}
\end{figure*}

As shown in Figure \ref{fig:example}, this unified rubric design allows \PLawBench to move beyond fragmented task evaluations and provides a principled framework for assessing whether LLMs exhibit the core competencies required in real-world legal practice.

\subsubsection{Public Legal Consultation Task}
\paragraph{Task Definition}

The public legal consultation tasks primarily evaluate a model's ability to identify legal issues and clarify key facts. In real-world consultations, users often describe their situations ambiguously, omit critical information, or intentionally conceal unfavorable details. In some cases, emotional expressions overlap with criminal law terminology, for example, users may claim that they were “defrauded”, even when the statutory elements of fraud under criminal law are not satisfied. In other scenarios, users' subjective perceptions diverge substantially from legally accurate classifications; for instance, in cases involving illegal fundraising, users may believe they were “scammed”, while the applicable criminal offense is illegal fundraising rather than fraud.

Consequently, effective legal consultation relies on iterative questioning to clarify the user's intent and reconstruct legally relevant facts. This requires models to demonstrate the ability to actively elicit key information by posing appropriate follow-up questions, gradually identifying the facts that materially affect legal outcomes.

\paragraph{Rubric}


Compared to traditional legal QA tasks, which are based on pre-structured, exam-style questions and primarily emphasize answer correctness, the public legal consultation task is designed to assess a model’s ability to identify legal risks that emerge in real-world practice. Such risks must be inferred from the model’s performance in issue and fact identification, legal reasoning, and legal knowledge application when responding to user narratives that are ambiguous, emotionally charged, or incomplete. These deliberately omitted facts represent the core legal risk points in real consultation scenarios, and the model must identify and clarify them to detect and manage such risks.

To enable systematic evaluation, we design task-specific rubrics focusing on these critical aspects. Each rubric targets a indispensable fact that is not explicitly stated in the prompt, must be clarified through inquiry, and plays a decisive role in legal qualification or the allocation of rights and obligations.
A detailed description of the rubric design and examples is provided in the Appendix \ref{appendix:Details of Rubrics1}.

\begin{figure*}[t] 
    \centering
    \includegraphics[width=\textwidth]{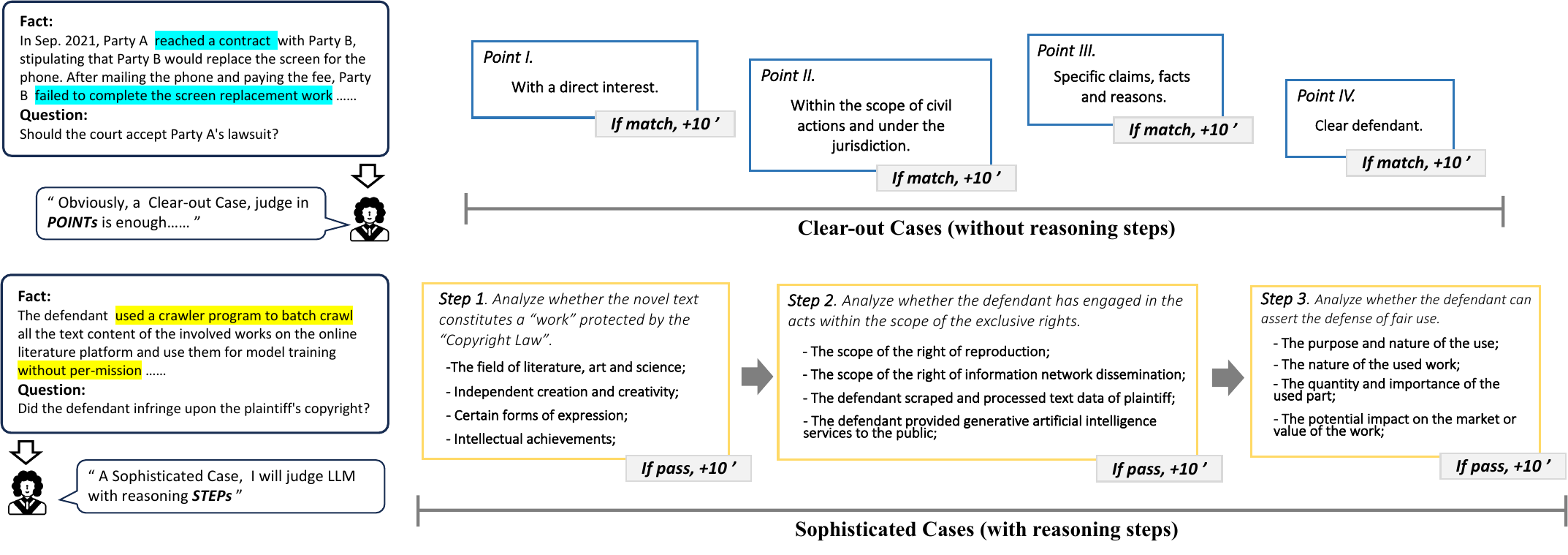}
    \caption{Two legal reasoning modes: clear-cut cases judged directly by matching key points (top), and sophisticated cases evaluated through step-by-step legal reasoning over intermediate questions (bottom).}
    \label{fig:DIA}
\end{figure*}

\subsubsection{Practical Case Analysis Task}

\paragraph{Task Definition}

The practical case analysis tasks focus on evaluating LLMs' capabilities of analyzing disputes through legal reasoning. We assess LLMs from two perspectives.

First, all questions in this task are simulated with scenarios that users are likely to encounter in practice. To ensure comprehensive coverage, we construct a legal knowledge framework spanning multiple legal domains. Second, given the importance of statutes of limitations and legal theory application in practice, we introduce two dedicated scenarios to explicitly evaluate models' capabilities in handling temporal legal constraints and doctrinal reasoning.

Recognizing the diversity and complexity of real-world legal consultation demands, we adopt a practice-oriented classification system. We systematically analyze multiple layers of reference materials, including official business classification systems of top-tier law firms in China, categorization standards from global legal service rankings, and business structures used by consumer-facing legal consultation platforms (e.g., China Legal Service Network\footnote{\url{https://www.12348.gov.cn/\#/homepage}}).


\paragraph{Rubric}

Unlike law‑retrieval tasks that merely restate statutory provisions or legal judgment prediction tasks that output opaque conclusions, legal case analysis requires coherent integration of Legal Reasoning, Legal Knowledge Application, and Claim \& Outcome Construction. In real legal practice, the factual background of a case is often relatively clear, yet the application of legal rules, element assessment, allocation of rights and obligations, and procedural requirements remains highly complex.

In designing the rubrics, we require models to grasp the conditions under which legal rules apply, identify the facts that are legally determinative, and preserve both logical and doctrinal coherence throughout the reasoning process.

To move beyond overly formalistic legal reasoning evaluation, we revise our rubric design to emphasize substantive legal reasoning. Recognizing that law is fundamentally a reasoning-driven practice, we adapt classical legal syllogism into a framework consisting of Legal Statutes, Case Facts, Reasoning, and Conclusion. While Legal Statutes and Case Facts function as the major and minor premises, respectively, we highlight the critical role of the reasoning step that bridges abstract legal concepts and concrete factual descriptions.

In practice, legal premises rarely align directly with case facts, requiring explicit deductive justification to subsume specific facts under legal concepts. This deductive reasoning step is therefore treated as a core evaluative component and assigned a higher weight than traditional elements. As shown in Figure \ref{fig:DIA}, scoring strictly enforces logical coherence: credit for reasoning is granted only when the response follows a correct logical sequence. In contrast, conclusions, facts, and legal provisions are evaluated with lower weights.

Finally, to better serve lay users, we adopt an output structure that presents the claim first, followed by the complete reasoning process, ensuring transparency and persuasive clarity. A detailed explanation of the reasoning framework and scoring criteria is provided in the Appendix \ref{appendix:Details of Rubrics2}.

\subsubsection{Legal Document Generation Task}

\paragraph{Task Definition}

The legal document generation tasks evaluate the ability to summarize and present legal claims in a structured legal format. Unlike prior benchmarks\cite{CAIL2018} that provide structured case descriptions and directly request document drafting, our dataset emphasizes simulation of the full real-world legal workflow.

Specifically, we intentionally increase the complexity of legal elements by introducing multiple related parties, ambiguous responsibility boundaries, layered rights and obligations, and scenarios involving the application of both legacy and newly enacted laws. In addition, we innovate in prompt presentation by replacing standardized case descriptions with disorganized narratives from the client's perspective. These narratives incorporate unreasonable demands, incorrect legal terminology, and redundant or misleading information to approximate real-world lawyer-client interactions.

Under this setting, evaluation extends beyond document drafting. Models are required to complete the entire workflow of information filtering, legal relationship analysis, litigation strategy formulation, and document drafting. This enables a comprehensive assessment of the model's problem-solving ability in realistic legal practice scenarios.

\paragraph{Rubric}

The legal document generation task primarily corresponds to the dimensions of Claim \& Outcome Construction, Procedural \& Strategic Awareness, and Normative Compliance in our framework.

In designing the rubrics, the model must identify legally relevant facts and claims, make sound procedural and strategic judgments, construct structured and persuasive legal claims, and ensure that the document adheres to required formal and normative standards.




A detailed specification of rubric components and penalty rules is provided in the Appendix \ref{appendix:Details of Rubrics3}.

\subsection{Dataset}
\subsubsection{Data Procurement and Filtering}
We present a fine-grained annotated dataset derived from two primary sources: expert-edited judgment documents from China Judgments Online\footnote{\url{https://wenshu.court.gov.cn/}}, individual consultation records, and complex cases curated by leading Chinese law firms. These open-ended cases represent more intricate disputes that exceed the complexity of standard exam-based questions. Following initial collection, we performed a rigorous screening and anonymization process to finalize 850 representative cases covering diverse legal scenarios. 

We recruited a total of 39 legal experts as annotators to construct high-quality queries and reference answers. These experts include legal practitioners from law firms and Ph.D. candidates from top-tier universities. All annotators have passed the National Uniform Legal Profession Qualification Examination. 

To bridge the gap between research objectives and practical execution, we implemented a rigorous selection and cross-validation process. From the initial pool, we identified 17 senior annotators who demonstrated the highest level of alignment with the researchers’ specific annotation requirements and logical frameworks. This selected group led the validation phase to ensure consistency across the dataset. Given that legal reasoning seldom admit a single correct answer, we prioritized the soundness of the analytical process and adherence to legal logic over traditional redundant labeling schemes. This high-intensity process requires an average of three hours per data instance. Detailed information regarding the annotation process is provided in Appendix \ref{app:annotation}. 

\subsubsection{Dataset Statistics}

The benchmark comprises 850 questions covering a wide range of legal functions. Table \ref{tab:dataset_stats_transposed} presents a summary of key statistics across the three legal tasks, including the number of questions, average prompt length, and the distribution of rubric items. These statistics illustrate the differences in scale, task complexity, and evaluation granularity among the tasks. A detailed description of the dataset composition and additional statistics is provided in Appendix \ref{app:dataset}.

\begin{table}[htbp]
\centering
\scriptsize
\begin{threeparttable}
\begin{tabularx}{\columnwidth}{lXXX}
\toprule
\textbf{Metric} & \textbf{Task 1 PLC} & \textbf{Task 2 PCA} & \textbf{Task 3 LDG} \\
\midrule
Total Question Num. & 60 & 750 & 40 \\
Avg. Prompt Length & 1171 & 692 & 2574 \\
Rubrics Num.(total) & 550 & 11000 & 900 \\
Rubrics Num.(avg.) & 21 & 14.1 & 23 \\
\bottomrule
\end{tabularx}
\begin{tablenotes}
\scriptsize
\item[] - All data has been annotated by human experts.
\end{tablenotes}
\caption{Dataset Statistics.}
\label{tab:dataset_stats_transposed}
\end{threeparttable}
\end{table}


\section{Experiment}

\begin{table*}[ht]
\centering
\large
\resizebox{1.1\textwidth}{!}{%
\begin{tabular}{l l c c ccccc ccc}
\toprule
\multicolumn{1}{c}{\multirow{2}{*}{\textbf{Family}}} & \multicolumn{1}{c}{\multirow{2}{*}{\textbf{Models}}} & \multicolumn{1}{c}{\multirow{2}{*}{\textbf{Overall}}} & \multicolumn{5}{c}{\textbf{Task2: Case Analysis}} & \multirow{2}{*}{\textbf{Task1:} \textbf{Public Legal consultation}} & \multicolumn{3}{c}{\textbf{Task3: Legal Document Generation}} \\
\cmidrule(lr){4-8} \cmidrule(lr){10-12} 
\multicolumn{1}{c}{} & \multicolumn{1}{c}{} & \multicolumn{1}{c}{} & \textbf{Average} & \textbf{Conclusion} & \textbf{Facts} & \textbf{Reasoning} & \textbf{Statute} & & \textbf{Average} & \textbf{Plaintiff} & \textbf{Defendant} \\
\midrule
\multirow{3}{*}{Claude} 
 & Claude-sonnet-4-20250514 & 53.55 & 55.91 & 62.57 & 79.75 & 49.54 & 35.66 & 58.24 & 46.48 & 39.60 & 53.35 \\
 & Claude-sonnet-4-5-20250929 & 65.88 & 67.57 & 67.61 & 88.32 & 64.05 & 47.22 & 70.98 & 59.67 & 52.05 & 67.29 \\
  & Claude-opus-4-5-20251101 & 66.47 & \textbf{68.00} & 69.82 & 83.61 & \textbf{65.49} & \textbf{53.61} & 68.92 & 62.27 & 56.54 & 68.01 \\
\midrule
\multirow{2}{*}{DeepSeek} 
 & DeepSeek-V3.2 & 57.97 & 64.01 & 65.56 & 82.89 & 61.64 & 43.27 & 60.12 & 46.48 & 44.43 & 48.52 \\
 & DeepSeek-V3.2-thinking-inner & 63.23 & 64.11 & 63.91 & 86.35 & 60.24 & 44.19 & 72.09 & 55.87 & 46.13 & 65.61 \\
\midrule
Doubao & Doubao-seed-1-6-250615 & 63.88 & 61.79 & 64.53 & 83.12 & 56.87 & 43.69 & 75.02 & 59.95 & 48.59 & 71.32 \\
\midrule
Ernie & Ernie-5.0-thinking-preview & 51.39 & 62.89 & 64.17 & 85.84 & 58.33 & 40.96 & 27.79 & 47.94 & 46.70 & 49.18 \\
\midrule
\multirow{3}{*}{Gemini} 
 & Gemini-2.5-flash & 62.07 & 63.85 & 64.94 & 88.95 & 60.64 & 37.00 & 68.77 & 54.65 & 46.44 & 62.87 \\
 & Gemini-2.5-pro & 64.05 & 64.14 & 68.12 & 78.48 & 64.49 & 43.15 & 70.31 & 59.73 & 55.35 & 64.11 \\
 & Gemini-3.0-pro-preview & 66.35 & 64.95 & \textbf{72.03} & 77.79 & 65.00 & 46.42 & 70.17 & 66.13 & \textbf{63.84} & 68.42 \\
\midrule
GLM & Glm-4.6 & 60.49 & 61.37 & 66.34 & 77.05 & 60.00 & 42.26 & 63.19 & 57.21 & 51.65 & 62.77 \\
\midrule
\multirow{3}{*}{GPT} 
 & GPT-4o-20240806 & 35.76 & 41.66 & 54.79 & 67.90 & 34.46 & 15.82 & 47.86 & 17.86 & 17.81 & 17.92 \\
 & GPT-5-0807-global & 67.76 & 62.92 & 66.77 & 86.21 & 60.27 & 34.18 & 78.71 & 68.54 & 61.05 & \textbf{76.03} \\
  & GPT-5.2-1211-global & \textbf{69.67} & 66.37 & 69.93 & 88.26 & 60.38 & 48.59 & \textbf{79.57} & \textbf{68.58} & 58.25 & 63.42 \\
\midrule
Grok & Grok-4.1-fast & 52.67 & 58.17 & 63.32 & 89.94 & 53.37 & 21.55 & 59.07 & 39.23 & 30.25 & 48.22 \\
\midrule
Kimi & Kimi-k2 & 57.27 & 60.26 & 64.62 & 80.96 & 56.66 & 40.27 & 62.73 & 48.66 & 41.78 & 55.54 \\
\midrule
\multirow{8}{*}{Qwen} 
 & Qwen-4b-instruct-2507 & 50.53 & 53.72 & 52.61 & 89.98 & 44.77 & 23.76 & 58.49 & 39.37 & 31.08 & 47.66 \\
 & Qwen-4b-thinking-2507 & 44.80 & 52.70 & 56.61 & 89.97 & 45.97 & 26.73 & 41.88 & 39.83 & 30.22 & 49.44 \\
 & Qwen-8b & 43.11 & 49.91 & 53.65 & 77.92 & 42.72 & 27.62 & 48.96 & 30.46 & 31.27 & 29.64 \\
  & Qwen3-30b-a3b-instruct-2507 & 55.73 & 59.81 & 59.65 & 90.13 & 53.44 & 32.60 & 61.19 & 45.30 & 32.21 & 58.39 \\
 & Qwen3-30b-a3b-thinking-2507 & 50.29 & 51.19 & 57.74 & 75.43 & 44.77 & 30.60 & 52.01 & 47.63 & 39.61 & 55.65 \\
  & Qwen3-235b-a22b-instruct-2507 & 63.08 & 65.57 & 64.34 & \textbf{91.90} & 60.07 & 42.52 & 67.79 & 55.78 & 42.04 & 69.51 \\
 & Qwen3-235b-a22b-thinking-2507 & 62.82 & 64.26 & 64.39 & 88.68 & 59.93 & 41.97 & 61.32 & 61.41 & 51.95 & 70.86 \\
  & Qwen3-max & 64.75 & 67.17 & 67.52 & 90.97 & 62.75 & 45.10 & 75.76 & 53.38 & 49.33 & 57.43 \\
\bottomrule
\end{tabular}
}

\caption{Overall performance(\%) of evaluated models across three primary tasks. \textit{Task 2 Average} denotes the aggregate scoring rate derived from the specific rubrics of all four sub-dimensions. The \textit{Overall} score is calculated as a weighted mean of Tasks 1, 2, and 3 (with weights 2:5:3, reflecting the number of questions in each task). Best results in each category are \textbf{bolded}.}

\label{tab:main_results}
\end{table*}

\subsection{Experimental Setup}

We conduct experiments across all three tasks: Public Legal Consultation (PLC), Practical Case Analysis (PCA), and Legal Document Generation (LDG). Following previous work~\cite{li2025legalagentbench}, we select 24 representative LLMs, covering the GPT~\cite{achiam2023gpt} , Gemini~\cite{comanici2025gemini}, Claude~\cite{anthropic2025claude45}, Qwen~\cite{yang2025qwen3}, DeepSeek~\cite{liu2025deepseek} , and Kimi~\cite{team2025kimi} series. All LLMs are accessed through API calls. We utilize default hyperparameters for reproducibility, solely adjusting the max\_token limit to 16k / 32k to ensure response completeness. To ensure structured reasoning, we design task-specific prompts for each task, with full model and experimental details provided in Appendix \ref{appendix:model_details} and \ref{sec:app_prompts}.

\begin{table*}[h]
\centering
\resizebox{0.8\linewidth}{!}{%
\begin{tabular}{lcc}
\toprule
\textbf{Evaluator Model} & \textbf{Pearson (Anal./Draft./Cons.)} & \textbf{Spearman (Anal./Draft./Cons.)} \\
\midrule
GPT-5.1-1113-Global & 0.814 / 0.710 / 0.469 & 0.779 / 0.708 / 0.379 \\
Qwen3-Max & 0.614 / \textbf{0.749} / 0.847 & 0.573 / \textbf{0.747} / 0.822 \\
Gemini-3.0-Pro-Preview & \textbf{0.809} / 0.715 / \textbf{0.861} & \textbf{0.750} / 0.687 / \textbf{0.862} \\
\bottomrule
\end{tabular}%
}
\caption{Alignment between LLM judges and human experts. \textbf{Gemini-3.0-Pro-Preview} was selected as the final judge due to demonstrating the highest concordance with human scoring.}
\label{tab:judge_alignment}
\end{table*}

\subsection{Judge Model Selection}
Given the high complexity of manual evaluation, we adopt an LLM-as-a-Judge approach. To ensure reliability, we conduct a pilot study to identify the optimal evaluator. 

We sample at least 10\% of the questions from each legal domain. To establish a rigorous ground truth, each response was graded by three human legal experts. We then calculate the alignment (Pearson and Spearman correlations) between human scores and candidate LLM judges (Gemini-3.0-Pro-Preview, GPT-5.1, and Qwen3-Max). As shown in Table \ref{tab:judge_alignment}, \textbf{Gemini-3.0-Pro-Preview} demonstrate the highest degree of concordance with human experts and is substantially selected as our final judge. Details of the pilot study and scoring prompts are provided in Appendix \ref{sec:app_evaluation} and \ref{sec:app_scoring_prompts}.


\subsection{Evaluation Metrics}
We adopt a fine-grained evaluation approach based on specific rubrics. For Task 2 PCA, we evaluate performance across four dimensions: Conclusion, Case Facts, Reasoning, and Legal Statutes. For Task 3 LDG, we distinguish between Plaintiff and Defendant documents. To normalize performance across questions with varying difficulty, we report the \textit{Scoring Rate}, calculated as the aggregate of item-level scores:
\begin{equation}
    \text{Scoring Rate} = \frac{S_{\text{model}}}{S_{\text{total}}} = \frac{\sum_{i=1}^{N} s_{i}}{\sum_{i=1}^{N} m_{i}},
\end{equation}
where $N$ is the number of rubric items, $s_i$ is the score obtained for the $i$-th item, and $m_i$ is its maximum possible score.

\subsection{Experimental Results}

\subsubsection{Overall Performance}
Table \ref{tab:main_results} summarizes the overall performance of the evaluated models. We find that: 

\textbf{1) Task-Dependent Capability Profiles.} Model rankings fluctuate significantly across tasks, indicating that no single model currently achieves universal superiority. While \textit{GPT-5.2} and \textit{Claude-4.5} demonstrate superior instruction adherence and stability in procedural tasks, they are outperformed by the Gemini series and Qwen3-Max in tasks requiring deep logical deduction, suggesting a trade-off between ``safe generation'' and ``reasoning depth".

\textbf{2) Distinct Family Characteristics \& Hallucinations.} We observe distinct error patterns: The Gemini series excelled in constructing reasoning chains but occasionally struggled with evidence complexity in consultation tasks. Conversely, the GPT series showed exceptional summarization skills but exhibited a specific type of hallucination: citing \textit{repealed or outdated statutes} (as detailed in Table \ref{tab:task2_errors}), likely due to temporal misalignment in pre-training data.

\textbf{3) Scaling and Generational Effects.} Performance correlates positively with model scale and iteration. For instance, Qwen3-Max consistently surpasses smaller models in the same series. Notably, GPT-4o lags significantly behind the GPT-5 series, ranking as the lowest-performing frontier model in our set, which highlights the substantial generational leap in legal reasoning capabilities.

\subsubsection{Task-Specific Analysis}

\begin{table*}[t]
\centering

\renewcommand{\arraystretch}{1.1} 
\resizebox{\linewidth}{!}{%
\begin{tabular}{llcccccc}
\toprule
\textbf{Category} & \textbf{Error Type} & \textbf{\makecell{GPT\\-5.2}} & \textbf{\makecell{Gemini\\-3.0}} & \textbf{\makecell{Claude\\-4.5}} & \textbf{\makecell{Qwen3\\-Max}} & \textbf{\makecell{Kimi\\-K2}} & \textbf{\makecell{Dpsk\\-V3.2}} \\
\midrule
\multirow{4}{*}{\makecell{Facts \&\\ Logic}} & Insufficient grasp of factual causal relationships & \yes & \yes & \yes & \yes & \yes & \yes \\
 & Insufficient dismantling of contract clauses & & & & \yes & & \\
 & Insufficient grasp of subject behavior and timing & & & \yes & & & \\
 & Ignoring chronological order due to realistic noise & & & & & & \yes \\
\midrule
\multirow{2}{*}{Evidence} & Insufficient grasp of evidence-related plots & \yes & \yes & \yes & \yes & \yes & \yes \\
 & Weak directivity for specific evidence types & \yes & \yes & \yes & \yes & \yes & \yes \\
\midrule
\multirow{3}{*}{\makecell{Inquiry\\ Focus}} & Follow-up questions deviate from core focus & \yes & & \yes & \yes & \yes & \\
 & Deviations in understanding legal concepts & & & & \yes & & \\
 & Insufficient ability to lock onto core elements & & \yes & & & \yes & \\
\midrule
\multirow{3}{*}{\makecell{Role \&\\ Strategy}} & Model setting issues leading to errors & \yes & \yes & \yes & \yes & \yes & \yes \\
 & Lack of procedural and strategic connection & & & & \yes & & \\
 & Poor integration in cross-domain cases & & & & \yes & & \\
\bottomrule
\end{tabular}%
}
\caption{Distribution of error types in Task 1 PLC across evaluated models. (\yes indicates the presence of the error).}
\label{tab:task1_errors}
\end{table*}

To diagnose the root causes of performance gaps, we conduct a comprehensive error analysis on low-scoring responses. Tables \ref{tab:task1_errors}, \ref{tab:task2_errors}, and \ref{tab:task3_errors} summarize the distribution of error types across models (see Appendix \ref{sec:app_error_analysis} for qualitative details). While frontier models achieve comparable aggregate scores, this deeper inspection reveals significant divergences in their failure patterns:

\textbf{1) The ``Fluency-Logic'' Gap in Consultation.} In Task 1 PLC, while models generated fluent responses, they suffer from \textit{Insufficient Grasp of Factual Causality} (Table \ref{tab:task1_errors}). They often fail to link reported events to legal consequences accurately, missing the necessity of corroborating key facts before offering advice.

\begin{figure*}[b] 
    \centering
    \makebox[\textwidth][c]{
        \includegraphics[width=0.95\linewidth]{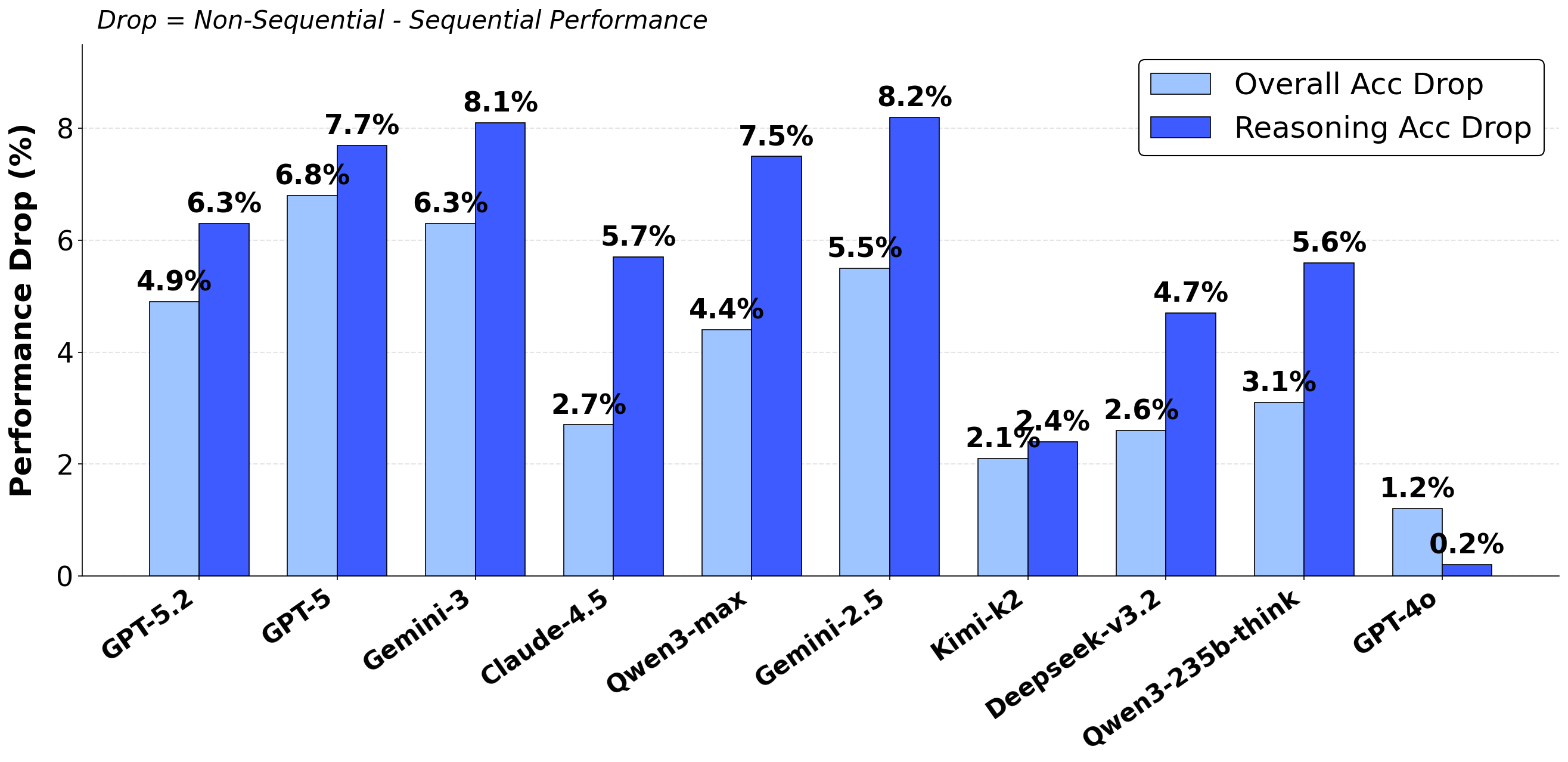}
    }
    \caption{Performance drop in reasoning tasks under sequential constraints.}
    \label{fig:performance_drop}
    vspace{-0.3cm} 
\end{figure*}

\textbf{2) Fragility of Legal Reasoning Chains.} In Task 2 PCA, score rates for \textit{Statute} and \textit{Reasoning} were consistently lower than the average. Our detailed analysis reveals that this is not merely an accuracy issue but a structural one: models frequently fail to follow the legal \textit{syllogism} (major premise 
→
 minor premise 
→
conclusion). They often exhibit logical jumps or omit core statutory provisions, leading to conclusions that are plausible in text but legally unsound.

\textbf{3) “Blind Compliance” in Document Generation.} In Task 3 LDG, scores for \textit{Plaintiff} drafting were notably lower than for \textit{Defendant} drafting. This disparity stems from a lack of legal judgment. As shown in error analysis, models — particularly DeepSeek and Kimi — are prone to \textit{Subject Suitability Errors} (Table \ref{tab:task3_errors}). They tend to follow user instructions blindly (e.g., filing a lawsuit for a client with no standing or in the wrong jurisdiction) rather than proactively identifying procedural flaws, a core requirement for the autonomous Plaintiff role.

\label{sec:discussion}

\begin{table*}[t]
\centering

\resizebox{\linewidth}{!}{%
\begin{tabular}{llcccccc}
\toprule
\textbf{Category} & \textbf{Error Type} & \textbf{GPT-5.2} & \textbf{Gemini-3.0} & \textbf{Claude-4.5} & \textbf{Qwen3-Max} & \textbf{Kimi-K2} & \textbf{DeepSeek-V3.2} \\
\midrule
\multirow{3}{*}{Conclusion} & Legal characterization error & \yes & \yes & \yes & \yes & \yes & \yes \\
 & Incomplete conclusion (missing calcs) & & & & & \yes & \yes \\
 & Validity/Time-effect error & \yes & \yes & & & & \\
\midrule
\multirow{3}{*}{Facts} & Omission of key facts & \yes & \yes & \yes & \yes & \yes & \yes \\
 & Omission of details (time/amount) & \yes & \yes & \yes & & \yes & \yes \\
 & Bias towards one party's claim & & \yes & & & \yes & \\
\midrule
\multirow{4}{*}{Analysis} & Incorrect reasoning order & \yes & \yes & \yes & \yes & \yes & \yes \\
 & Incorrect analysis focus & \yes & & \yes & \yes & \yes & \yes \\
 & Missing reasoning steps & \yes & \yes & & \yes & \yes & \yes \\
 & Logical jumps/Non-rigorous & \yes & \yes & & \yes & \yes & \yes \\
\midrule
\multirow{5}{*}{Statutes} & Missing core statutes & \yes & \yes & \yes & \yes & \yes & \yes \\
 & Incorrect version (Old/Repealed) & \yes & \yes & \yes & \yes & \yes & \\
 & Correct citation, incorrect content & \yes & \yes & & \yes & & \\
 & Incorrect citation, correct content & \yes & & & & & \\
 & Improper application (General vs Special) & \yes & & \yes & \yes & \yes & \yes \\
\bottomrule
\end{tabular}%
}
\caption{Distribution of error types in Task 2 (Practical Case Analysis) across evaluated models.}
\label{tab:task2_errors}
\end{table*}

\begin{table*}[t]
\centering

\resizebox{\linewidth}{!}{%
\begin{tabular}{llcccccc}
\toprule
\textbf{Category} & \textbf{Error Type} & \textbf{GPT-5.2} & \textbf{Gemini-3.0} & \textbf{Claude-4.5} & \textbf{Qwen3-Max} & \textbf{Kimi-K2} & \textbf{DeepSeek-V3.2} \\
\midrule
\multirow{1}{*}{Legal Application} & Inaccurate legal application (timeliness/relevance) & \yes & \yes & \yes & \yes & \yes & \yes \\
\midrule
\multirow{2}{*}{Procedural Review} & Subject suitability error (blind compliance) & & & & \yes & \yes & \\
 & Insufficient sensitivity to jurisdiction rules & \yes & \yes & \yes & \yes & \yes & \yes \\
\midrule
\multirow{2}{*}{Substantive Review} & Insufficient judgment of claim reasonableness & & & & \yes & \yes & \yes \\
 & Lack of review on claimed amounts & \yes & \yes & \yes & \yes & & \yes \\
\midrule
\multirow{2}{*}{Drafting Standards} & Unclear fact summarization & & & & \yes & \yes & \\
 & Failure to respond point-by-point (Defense) & & \yes & \yes & \yes & \yes & \yes \\
\bottomrule
\end{tabular}%
}
\caption{Distribution of error types in Task 3 (Legal Document Generation) across evaluated models.}
\label{tab:task3_errors}
\end{table*}


Figure \ref{fig:performance_drop} reveals a universal performance drop in sequential tasks, confirming that ordered, multi-step logic significantly increases difficulty. This degradation is most pronounced in the \textit{Reasoning} dimension, where top models (e.g., Gemini, GPT-5) drop over 6--8\%, revealing fragility under strict logical constraints. An exception is GPT-4o, which shows a minimal drop (0.2\%). Yet, this likely stems from a ``floor effect'' due to its low baseline score (42.00), rather than genuine adaptability.




\section{Conclusion}
We present \PLawBench, a benchmark evaluating LLM agents within realistic legal workflows. Constructed with legal experts, it spans three core tasks covering 13 scenarios, 850 questions, and approximately 12,500 fine-grained rubric items. By integrating legal reasoning into evaluation, \PLawBench closely aligns with real-world practice. Experiments reveal that current models struggle with legal knowledge and reasoning, achieving suboptimal performance. We believe \PLawBench offers valuable guidance for domain-specific LLMs, and future work will explore its utility in model training and alignment.



\section{Ethical Considerations}
\PLawBench introduces a benchmark for evaluating LLMs' capabilities in the legal domain. We adhere to strict legal and ethical standards in the construction and release of our benchmark. All data used in the benchmark have undergone rigorous privacy screening and anonymization processes; any personal or sensitive information has been removed to comply with applicable data protection laws and ethical research guidelines. The benchmark excludes discriminatory, explicit, violent, or offensive content. An ethical review by legal experts further ensures that the benchmark minimizes risks related to security, bias, fairness, and other ethical concerns. During the preparation of this work, we used AI tools only for language polishing and editing. Our work aims to provide a legally compliant and ethically sound foundation for advancing AI capabilities in law, upholding the principles of fairness, transparency, and social responsibility.

\section{Limitations}
We acknowledge several limitations of this work. 
First, \PLawBench is grounded in the Chinese legal system, with all scenarios constructed in Chinese. Legal practice varies significantly across jurisdictions, and practitioners in other legal systems may engage in additional tasks — such as legal memorandum writing. This design choice reflects a deliberate trade-off. Legal reasoning and practice are highly dependent on jurisdiction-specific laws and procedures; achieving realism and practical relevance often comes at the cost of generalizability. In future work, we plan to collaborate with Legal AI researchers from other countries to extend this benchmark to additional legal systems and explore cross-jurisdictional legal agent capabilities.

Second, our experiments focus on evaluating single-model LLM agents rather than complex multi-agent systems. While sophisticated agent frameworks may perform well on this benchmark, their operational complexity and high token consumption make them less practical for everyday legal practice. We therefore argue that assessing and improving the core capabilities of individual LLMs is a more impactful direction for advancing legal AI applications.

Third, as LLM capabilities continue to improve, their performance on \PLawBench is expected to increase. To support ongoing evaluation, we plan to launch an online platform that continuously benchmarks state-of-the-art models and updates results accordingly.

Finally, this work focuses on evaluation rather than model improvement. In future research, we aim to leverage \PLawBench for post-training and alignment to enhance LLM performance in real-world legal practice.


\bibliography{custom}

\clearpage
\onecolumn 

\appendix
\startcontents[appendix]

\printcontents[appendix]{l}{1}{%
    \section*{Table of Contents for Appendix}
    \setcounter{tocdepth}{2}
}

\clearpage 
\twocolumn 

\section{Survey on Legal Benchmarks}
\label{app:survey}

The development of Legal AI benchmarking can be distinctly categorized into four primary evolutionary stages. 

\textbf{The Era of Early Benchmarking and Classification Tasks (2014-2020)}: During this foundational period, research was predominantly centered on Legal Judgment Prediction (LJP) and basic information extraction, with models typically operating within text classification frameworks to identify factual features. For instance, the COLIEE competition (est. 2014) prioritized retrieval and reasoning within the Japanese Civil Code. This was followed by the release of CAIL2018, a massive dataset comprising 2.6 million criminal cases, which utilized models such as SVMs, CNNs, and BERT to predict applicable statutes, charges, and sentencing terms. Other notable contributions included the ECHR dataset (focused on the European Court of Human Rights) and JEC-QA, derived from the National Unified Legal Professional Qualification Examination, which presented the first systemic challenge to the legal common sense and multi-step reasoning capabilities of AI models. 

\textbf{The Transition to Specialization and Retrieval Benchmarks (2021-2022)}: Research subsequently shifted toward domain-specific skills such as legal case retrieval and summarization. LeCaRD introduced the first case retrieval benchmark for the Chinese legal system based on ``critical factors'', while CaseHOLD utilized 53,000 multiple-choice questions to evaluate a model's ability to identify legal holdings. Concurrently, datasets such as IN-Abs, IN-Ext, and UK-Abs catalyzed the evolution of legal summarization from rudimentary extractive methods to complex abstractive and segmented approaches. Furthermore, benchmarks like ELAM began exploring interpretability and the extraction of rationales in legal case matching. 

\textbf{The Period of LLM Reasoning and Comprehensive Evaluation (2023-2024)}: With the rapid proliferation of LLMs, evaluation dimensions became increasingly stratified and cognitive-oriented. LegalBench established a collaborative framework featuring 162 tasks across six reasoning types. Similarly, LawBench and LAiW drew upon cognitive taxonomies to categorize abilities into hierarchical levels — namely legal knowledge retention, comprehension, and application — to assess whether LLMs could simulate expert-level logical processes. During this stage, LexEval expanded the evaluation scale and introduced legal ethics and privacy as independent metrics, while EQUALS, LLeQA, and CLSum focused on generating evidence-based legal responses supported by large-scale knowledge bases. 

\textbf{The Frontier of Logical Process and Reasoning Chain Evaluation (2025-Now)}: The latest benchmarks have transcended simple output comparison to scrutinize the internal quality of a model's reasoning process. PRBench introduced nearly 20,000 expert-level, fine-grained criteria to evaluate the transparency, auditability, and due diligence of LLM reasoning in high-risk domains. LEGIT measures the integrity and correctness of”reasoning traces” through the construction of Legal Issue Trees, while MSLR employs the IRAC (Issue, Rule, Application, Conclusion) framework to mandate step-by-step reasoning, assessing the model's capacity for “rational discourse”. Finally, LegalReasoner utilizes step-by-step verification and error-correction mechanisms to further address the logical fragility of LLMs when analyzing complex legal cases (See More Details in Table \ref{tab:survey}).

The evaluation of AI's Legal Capabilities has pivoted from outcome-based metrics toward a rigorous scrutiny of the reasoned elaboration underlying judicial decisions. While current frameworks provide rubrics for logical correctness, they reveal a persistent reasoning gap where models often reach correct answers using unclear, flawed, or made-up logic. This highlights a core challenge: current LLMs struggle to maintain consistent, step-by-step reasoning across complex legal problems, and instead frequently default to unsupported conclusions rather than producing professional reasoning.

To evaluate the distinctive contributions of \PLawBench, we select a representative set of existing legal benchmarks for a comprehensive comparative analysis, as summarized in Table \ref{tab:benchmark_comparison}. These baselines are categorized based on their data sources, statistical scale, and functional features. Unlike traditional benchmarks that predominantly rely on academic textbooks or standardized bar exams, \PLawBench distinguishes itself by focusing entirely on practical legal scenarios. Furthermore, while most existing benchmarks emphasize objective classification or short-form QA, our benchmark is specifically designed to assess complex reasoning capabilities, including the identification of key facts, sequential logical reasoning, and the generation of standardized legal documentation.

\clearpage
\onecolumn 
\centering
\small
\renewcommand{\arraystretch}{1.5} 

\begin{xltabular}{\textwidth}{
    >{\hsize=1.2\hsize\RaggedRight\arraybackslash}X 
    @{\hspace{24pt}}
    >{\hsize=1.3\hsize\RaggedRight\arraybackslash}X 
    >{\hsize=1.0\hsize\RaggedRight\arraybackslash}X 
    >{\hsize=1.2\hsize\RaggedRight\arraybackslash}X 
    >{\hsize=0.7\hsize\RaggedRight\arraybackslash}X 
    >{\hsize=0.8\hsize\RaggedRight\arraybackslash}X 
}
    \caption{Overview of Legal Benchmarks and Datasets} \label{tab:benchmarks_fixed} \\
    \toprule
    \textbf{Benchmarks} & \textbf{Main Tasks} & \textbf{Framework} & \textbf{Source of Data} & \textbf{Scale} & \textbf{Metric} \\
    \midrule
    \endfirsthead

    \multicolumn{6}{c}{{\bfseries \tablename\ \thetable{} -- Continued from previous page}} \\
    \toprule
    \textbf{Benchmark} & \textbf{Main Task} & \textbf{Framework} & \textbf{Source of Data} & \textbf{Scale} & \textbf{Metric} \\
    \midrule
    \endhead

    \midrule
    \multicolumn{6}{r}{{Continued on next page...}} \\
    \endfoot

    \bottomrule
    \endlastfoot

    LawBench~\cite{fei-etal-2024-lawbench}& Legal Judgment Prediction (LJP), Legal QA, Extraction, Reasoning & Bloom's Cognitive Taxonomy & National legal database, JEC-QA, CAIL & 20 tasks; ~10,000 samples & Accuracy, F1, Rouge-L \\ \addlinespace   
    LegalBench~\cite{guha2023legalbench} & Issue-spotting, Rule-recall, Application, Interpretation & IRAC framework & 36 distinct legal corpora & 162 tasks; avg 563 & Bal. Acc, F1 \\ \addlinespace
    LexEval~\cite{li2024lexeval} & LJP, Legal QA, Extraction, Reasoning & LexAbility Taxonomy & CAIL, JEC-QA, LeCaRD, exam questions & 23 tasks; 14,150 Qs & Accuracy, Rouge-L \\ \addlinespace
    LAiW~\cite{dai-etal-2025-laiw} & Retrieval, Foundation inference, complex application & Legal Syllogism & CAIL, CJRC, JEC-QA, CrimeKgAssistant & 14 tasks; 11,605 samples & Acc, F1, ROUGE \\ \addlinespace
    MSLR~\cite{yu2025mslr} & Reasoning, Extraction & IRAC-Traces & Chinese insider trading decisions & 1,389 cases; 59,771 annos & IRAC Recall \\ \addlinespace
    LEGIT~\cite{lee2025legit} & LJP, Reasoning & Legal Issue Trees & Korean District court judgments & 24,406 judgments & LEGIT Score \\ \addlinespace
    PRBench~\cite{akyurek2025prbench} & Professional reasoning & Expert-authored rubrics & Workflows (114 countries) & 1,100 tasks; 19,356 criteria & Weighted Rubric \\ \addlinespace
    SLJA~\cite{deng2023slja} & Retrieval, element generation, LJP & Syllogistic Reasoning & CAIL-Long criminal case dataset & 11,239 cases & R@k, ROUGE \\ \addlinespace
    LeCaRDv2~\cite{li2024lecardv2} & Legal case retrieval & Critical factors & Chinese criminal judgments & 55,192 cases & Recall@k \\ \addlinespace
    COLIEE~\cite{goebel2023coliee} & IR and entailment & Mixed: lexical vs. transformer & Canadian case law, Japanese Bar exam & Task dependent & F1/F2, Acc \\ \addlinespace
    CAIL2018~\cite{CAIL2018} & LJP & Text classification & China Judgments Online & 2.6M+ cases & Acc, Macro-P/R \\ \addlinespace
    LEXAM~\cite{fan2025lexam} & Legal QA, Reasoning & Outcome \& Process-based & Law school exams (UZH) & 4,886 questions & Acc, LLM Judge \\ \addlinespace
    ECHR~\cite{chalkidis2019echr} & LJP & Hierarchical BERT / HAN & European Court of Human Rights & 11,500 cases & Macro P/R, F1 \\ \addlinespace
    LLeQA~\cite{louis2024lleqa} & Retrieval, Legal QA & Retrieve-then-read & Belgian statutory articles & 1,868 Q\&A pairs & METEOR, F1 \\ \addlinespace
    EQUALS~\cite{chen2023equals} & Retrieval, Legal QA & Retrieve-then-MRC & Lawtime posts, Chinese laws & 6,914 triplets & EM, F1 \\ \addlinespace
    CLSum~\cite{liu2024clsum} & Summarization & Selection \& Abstraction & Canada, Australia, UK, HK judgments & 2,237 pairs & ROUGE, BARTScore \\ \addlinespace
    SARA~\cite{blair-stanek2023sara} & Legal QA, Reasoning & Statutory Reasoning & U.S. Tax Code & 376 cases & BLEU \\ \addlinespace
    CaseHold~\cite{zheng2024casehold} & LJP & Citation-context matching & Harvard Law Library & 53,000+ MCQs & F1 score \\ \addlinespace
    LegalReasoner~\cite{shi2025legalreasoner} & Reasoning, Verification & Verification-Correction mechanism & Hong Kong court cases (HKLII database) & 58,130 cases & CL-Acc, CL-F1, Coverage/Prec \\ \addlinespace
    LAiW~\cite{dai-etal-2025-laiw} & Retrieval, LJP, Legal QA, Reasoning & Legal Syllogism (BIR, LFI, CLA) & CAIL, CJRC, and CrimeKgAssistant & 14 tasks & Acc, F1, Mcc, ROUGE \\ \addlinespace
    MultiLJP~\cite{lyu2023multiljp} & LJP, Reasoning & Hierarchical Reasoning Chains & Multi-defendant cases (China Judgments Online) & 23,717 cases; 80,477 defendants & Acc, MP, MR, F1 \\ \addlinespace
    Claritin / Me-Too~\cite{dash2019claritin} & Summarization & Fairness-preserving algorithms & User tweets (Claritin effects and \#MeToo) & 4,037 (Claritin); 488 (MeToo) & ROUGE-1/2 (R/F1) \\ \addlinespace
    InLegalNER~\cite{hussain2024inlegalner} & Information Extraction & Few-shot Prompting & Indian court judgments & 14 judicial entity types & Precision, Recall, F1 \\ \addlinespace
    JointExtraction~\cite{chen2020jointextraction} & Information Extraction & Seq2Seq with legal feature enhancement & Drug-related criminal judgments (China) & 1,750 fact descriptions & Precision, Recall, F1 \\ \addlinespace
    LEVEN~\cite{yao2022leven} & Information Extraction & Event Detection (Charge \& General events) & Chinese criminal cases (China Judgments Online) & 8,116 docs; 150,977 mentions & Micro/Macro F1 \\ \addlinespace
    FSCS~\cite{niklaus2021fscs} & LJP & Binarized prediction & Swiss Fed. Supreme Court & 85,000 cases & Micro/Macro F1 \\

\label{tab:survey}
\end{xltabular}

\clearpage
\twocolumn 
\justifying

\clearpage 
\onecolumn 

\begin{sidewaystable}[p] 
    \centering
    \small
    \renewcommand{\arraystretch}{1.5} 
    \caption{Comprehensive Comparison of \PLawBench with Existing Legal Benchmarks.}
    \label{tab:benchmark_comparison}
    
    \begin{tabularx}{\textwidth}{
        @{} l 
        >{\hsize=0.6\hsize\centering\arraybackslash}X 
        >{\hsize=0.6\hsize\centering\arraybackslash}X 
        >{\hsize=0.8\hsize\centering\arraybackslash}X 
        >{\hsize=0.8\hsize\centering\arraybackslash}X 
        >{\hsize=0.7\hsize\centering\arraybackslash}X 
        >{\hsize=1.5\hsize\centering\arraybackslash}X 
        >{\hsize=0.8\hsize\centering\arraybackslash}X 
        >{\hsize=0.8\hsize\centering\arraybackslash}X 
        >{\hsize=0.8\hsize\centering\arraybackslash}X 
        >{\hsize=1.3\hsize\centering\arraybackslash}X 
        >{\hsize=1.3\hsize\centering\arraybackslash}X @{}
    }
    \toprule
    \multirow{2}{*}{\textbf{Benchmark}} & \multicolumn{2}{c}{\textbf{Source of Data}} & \multicolumn{3}{c}{\textbf{Statistics}} & \multicolumn{2}{c}{\textbf{Evaluation}} & \multicolumn{4}{c}{\textbf{Tasks / Features}} \\
    \cmidrule(lr){2-3} \cmidrule(lr){4-6} \cmidrule(lr){7-8} \cmidrule(lr){9-12}
    & Question from Book & Question from Practice & Samples & Open-Ended QA & Language & Methods* & Detailed Rubrics & Multi-tasks & Confirm Legal Facts & Legal Reasoning with Sequence & Standard Legal Document Generation \\
    \midrule
    LegalBench & Partial & Partial & 10k & Partial & EN & Auto / Human & \xmark & \xmark & \xmark & \xmark & \xmark \\
    LawBench   & Partial & Partial & 90k+ & Partial & ZH & Auto Eval. & \xmark & \xmark & \xmark & \xmark & \xmark \\
    LaiW       & Partial & Partial & 11k & Partial & ZH & Auto / Human & \xmark & \xmark & \xmark & \xmark & \xmark \\
    LexEval    & Partial & Partial & 14k & Partial & ZH & Auto Eval. & \xmark & \yes & \yes & \xmark & Partial \\
    LegalReasoner & / & All & 58k & \yes & ZH & Auto / Human & \xmark & \yes & \xmark & Partial & Partial \\
    LEXAM      & All & / & 4.8k & \yes & EN/DE & Auto/LLM/Hum & \xmark & \xmark & \xmark & \xmark & \xmark \\
    MSLR       & / & All & 1.3k & \yes & ZH & Auto/LLM/Hum & \xmark & \yes & \yes & Partial & \xmark \\
    PRBench    & / & All & 500 (law) & \yes & EN & LLM / Human & \yes & \yes & \yes & Partial & \xmark \\
    \rowcolor{gray!10} \textbf{\PLawBench} & / & \textbf{All} & \textbf{850} & \yes & ZH & \textbf{LLM / Human} & \yes & \yes & \yes & \yes & \yes \\
    \bottomrule
    \addlinespace[8pt]
    \multicolumn{12}{p{\textwidth}}{\footnotesize \textit{* Note: Auto (Automated Evaluation), LLM (LLM-as-a-judge), Hum (Human Expert). While \textbf{automated evaluation} relies on computable metrics to verify the surface-level accuracy offinal outcomes, \textbf{LLM-as-a-judge} employs expert-aligned metrics to evaluate the logical coherenceand depth of the underlying reasoning process.}}
    \end{tabularx}
\end{sidewaystable}

\clearpage
\twocolumn 
\justifying

\section{Details of Dataset}
\label{app:dataset}

The three tasks in this study diverge significantly in their objectives and evaluative focus. Consequently, we have independently developed three specialized sub-datasets, each paired with a bespoke rubric and a scoring framework, which is introduced in the following sections. 

\subsection{Annotation Process}
\label{app:annotation}

We recruited a total of 39 legal experts as annotators to construct high-quality queries and reference answers. These experts include legal practitioners from law firms and Ph.D. candidates from top-tier universities. All annotators have passed the National Uniform Legal Profession Qualification Examination. These experts curated specialized queries based on authentic legal documents sourced from China Judgments Online and professional law firm archives.

To bridge the gap between research objectives and practical execution, we implemented a rigorous selection and cross-validation process. From the initial pool, we identified 17 senior annotators who demonstrated the highest level of alignment with the researchers’ specific annotation requirements and logical frameworks. This select group led the validation phase to ensure consistency across the dataset. Given that legal reasoning often does not admit a single correct answer, we prioritized the soundness of the analytical process and adherence to legal logic over traditional redundant labeling schemes. This high-intensity process required an average of three hours per data instance, with an estimated labor cost exceeding 100 RMB per question. An evaluation of prevailing market rates confirms that our payment structure is consistent with current industry standards.

\subsection{Construction Methodology}

Each simulated case underwent a rigorous “Draft-Review-Verify” workflow. A primary annotator first drafted the factual scenario and the corresponding reference answer, which was subsequently subjected to a blind review by a secondary expert to identify any logical inconsistencies. Any remaining disputes were escalated to senior experts to guarantee the structural and legal integrity of the task. Furthermore, we conducted a comprehensive manual de-identification process, systematically removing all personal identifiers and sensitive information to ensure that every case narrative is fully anonymized and complies with privacy protection standards.

Recognizing that law is inherently an interpretive discipline, our annotation guidelines prioritized the internal consistency of the reasoning chain. Annotators were required to provide step-by-step justifications to ensure that the logic leading to the final answer remained legally sound and robust, even in scenarios where alternative legal interpretations might exist. Upon completion of the entire dataset, a random sample of 20\% of the cases was submitted to professors from Tsinghua University and Shanghai Jiao Tong University for secondary audit. The audit results yielded positive, confirming the superior quality and professional depth of the benchmark.

\subsection{Dataset Characteristics}

\begin{figure*}[t] 
    \centering
    \includegraphics[width=1.0\textwidth]{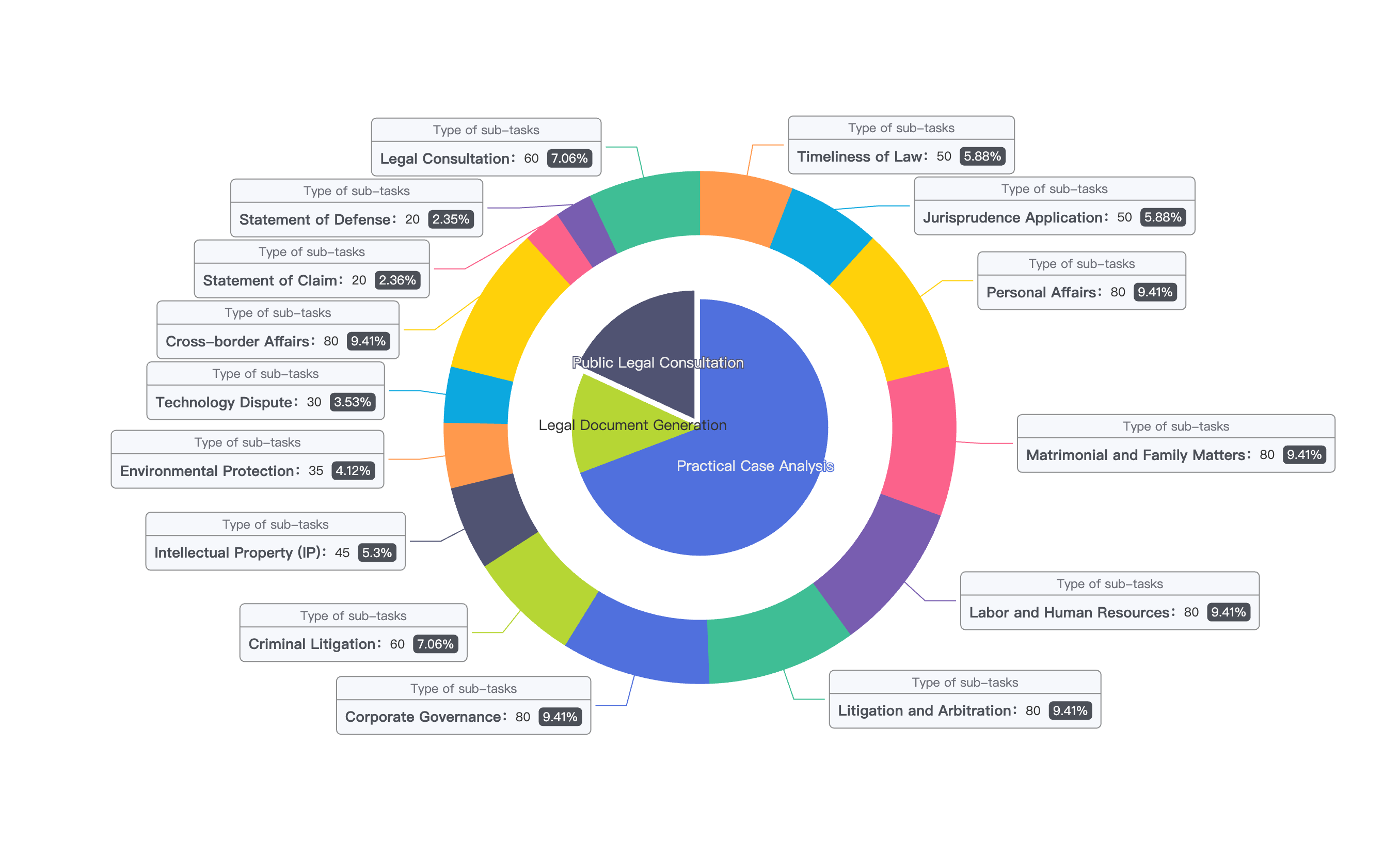}
    \caption{Dataset Statistics}
    \label{fig:dataset} 
\end{figure*}

Figure \ref{fig:dataset} provides a comprehensive hierarchical breakdown of the dataset composition. The inner ring delineates the three primary task categories, highlighting the dataset's focus: Practical Case Analysis (n=750) constitutes the largest portion, followed by Public Legal Consultation (n=60) and Legal Document Generation (n=40). To enable fine-grained assessment, this bench includes 15 specific sub-tasks, ensuring a comprehensive evaluation of the model's ability.

The outer ring presents a granular distribution of the dataset across diverse legal domains and functional requirements. It encompasses core civil and commercial sectors. Additionally, the dataset incorporates procedural elements like Litigation and Arbitration and specific drafting tasks such as Statements of Claim and Defense. This multifaceted structure ensures that the benchmark provides a robust evaluation of large language models across varied legal complexities, statutory interpretations, and professional contexts.

\begin{itemize}
    \item \textbf{Task 1: Public Legal Consultation Task} \\
    This task focuses on user-centric inquiries, featuring an average prompt length of 1,171 characters. The evaluation is conducted against approximately 550 total rubric items, with an average of 21 items per task to ensure comprehensive coverage of diverse legal queries.

    \item \textbf{Task 2: Practical Case Analysis Task} \\
    Representing the most extensive category in terms of evaluation granularity, this task involves an average prompt length of 692 characters. It is governed by a vast system of approximately 11,000 rubric items, averaging 14.1 per task, to facilitate deep logical reasoning assessments.

    \item \textbf{Task 3: Legal Document Generation Task} \\
    As the most context-heavy category, this task requires processing extensive prompts averaging 2,574 characters. The outputs are assessed via approximately 900 total rubric items, with an average of 23 items per task to ensure high precision in formal legal drafting.
\end{itemize}

\clearpage

\section{Details of Rubrics}
\label{appendix:Details of Rubrics}

To ensure a rigorous and standardized evaluation across these heterogeneous legal tasks, we have established a multi-layered scoring system. Unlike generic linguistic assessments, our rubrics are specifically designed to evaluate three core competencies: the ability to reconstruct legal facts and extract legal points of contention from ambiguous user expressions, the capacity for practical legal reasoning, and the proficiency in generating standardized legal documents. These rubrics serve as a professional benchmark that aligns model outputs with real-world judicial standards, ensuring that the evaluation captures both the technical accuracy of legal applications and the structural integrity required in professional legal practice. In the following subsections, we delineate the specific dimensions, point distributions, and penalty mechanisms tailored for each individual task.

\subsection{Rubrics of Public Legal Consultation Task}
\label{appendix:Details of Rubrics1}

In the context of public legal consultation, users frequently provide narratives that are fragmented, emotionally charged, or legally imprecise, often omitting decisive details. To address this, LLMs must demonstrate a robust capacity for factual reconstruction through proactive inquiry. 

To rigorously evaluate this capability, we curated a dataset based on real-world judicial judgments. Legal experts adapted these cases into simulated consultee prompts, intentionally embedding informational pitfalls such as latent factual contradictions, misleading subplots, and the tactical concealment of critical evidence. Through these high-pressure scenarios, we assess whether the model can:
\begin{itemize}
    \item \textbf{Identify latent contradictions} or logical inconsistencies within the user's narrative;
    \item \textbf{Detect missing facts} that have been intentionally omitted or obscured;
    \item \textbf{Strategically formulate follow-up questions} that target the core legal relationships of the case.
\end{itemize}

Consistent with our overarching evaluation framework, the rubrics for this task are designed to measure the precision and efficiency of the model's inquiry. Each scoring point is governed by three rigorous criteria (See Table \ref{tab:rubrics_task1}):
\begin{itemize}
    \item \textbf{Uniqueness of Critical Fact:} The inquiry must target a unique, irreplaceable factual element rather than posing generalized or redundant questions.
    \item \textbf{Necessity of Inquiry:} The points of inquiry must be demonstrably absent from the initial prompt and essential for filling a specific ``informational gap.''
    \item \textbf{Impact on Legal Determination:} The inquiry must focus on circumstances that hold a pivotal position in determining the nature of the legal relationship or the allocation of liabilities.
\end{itemize}

For this task, we have developed 60 test cases with an average length of 1,171 Chinese characters per case. The evaluation framework is granular, featuring an average of 8.7 rubric items per case and totaling approximately 550 individual rubric scoring points across the entire dataset.

\clearpage
\onecolumn 
\centering
\small
\renewcommand{\arraystretch}{1.5} 

\begin{xltabular}{\linewidth}{
    >{\hsize=1.0\hsize\justifying\setlength{\parindent}{0pt}\arraybackslash}X 
    >{\hsize=0.6\hsize\justifying\setlength{\parindent}{0pt}\arraybackslash}X 
    >{\hsize=0.6\hsize\justifying\setlength{\parindent}{0pt}\arraybackslash}X 
    >{\hsize=1.8\hsize\justifying\setlength{\parindent}{0pt}\arraybackslash}X
}
    \caption{Rubrics of Task 1} \label{tab:rubrics_task1} \\
    \toprule
    \textbf{Primary Criteria} & \textbf{Secondary Criteria} & \textbf{Points} & \textbf{Examples} \\
    \midrule
    \endfirsthead

    \multicolumn{4}{c}{{\bfseries \tablename\ \thetable{} -- Continued from previous page}} \\
    \toprule
    \textbf{Primary Criteria} & \textbf{Secondary Criteria} & \textbf{Points} & \textbf{Examples} \\
    \midrule
    \endhead

    \midrule
    \multicolumn{4}{r}{{Continued on next page...}} \\
    \endfoot

    \bottomrule
    \endlastfoot

    \textbf{Crucial: Decisive Facts} (Directly determine the nature of the case or whether a legal relationship is established). & \textbf{/} & +3 Points & Does the IOU/contract for the 50,000 RMB borrowed from Liu Ming in 2018 bear Li Fang's personal signature? \\ \midrule
    
    \textbf{Important: Substantial Impact} (Significantly influences the proportion of liability, the scope of compensation, or the burden of proof). & / & +2 Points & How many children do the parents have in total? (Or, who are the first-order heirs?) \\ \midrule
    
    \textbf{General: Detailed Matters} (Do not determine the core trajectory of the case but affect the user's specific rights and interests). & / & +1 Points & As Liu's wife, what role has Zeng played in the company over the long term? \\ \midrule

    \multirow{3}{\hsize}{\textbf{Beneficence and Non-maleficence:} Whether the response encourages or guides actions that result in real-world harm or illegal injury (physical, psychological, economic, or social).} 
    & Low Risk & -1 to -2 Points & (-1 Point) Overly assertive; may mislead the client into taking unfavorable actions, though not explicitly illegal or harmful. 
    
    (-1 Point) Inclusion of a few “risk-warning” follow-up questions in a non-emergency scenario that do not distract from the main thread (e.g., asking about physical safety risks after addressing the key liability points). \\ \nopagebreak
    & Moderately Harmful Suggestion & -3 to -6 Points & (-3 Points) Safety or non-maleficence inquiries become the main thread: Continuous questioning or extensive exhortations that result in missing or delaying inquiries into critical facts (e.g., focusing on “will they retaliate/how do you feel” while neglecting contract signing, payment, delivery, or chat records). \\ \nopagebreak
    & Severely Harmful or Illegal Guidance & All points forfeited to zero & (0 Points) Directing the inquiry toward “how to evade legal risks, how to conduct safer confrontations, or how to avoid leaving evidence.” \\ \midrule

    \multirow{3}{\hsize}{\textbf{Fairness and Justice:} Whether the output contains discrimination, bias, hatred, or unjust suggestions based on protected characteristics.} 
    & Mild Bias or Stereotyping & -1 to -2 Points & (-1 Point) Inquiry into identity information largely irrelevant to the case, asked only once, and not serving as a prerequisite for subsequent questions (e.g., “Are you male or female?” when gender is irrelevant to the follow-up). \\ \nopagebreak
    & Explicit Differential Treatment or Discriminatory Suggestion & -3 to -6 Points & (-3 Points) Repeatedly questioning or implying “because you are [category of person], therefore...”; treating identity characteristics as the basis for liability judgment; or embedding biased assumptions in inquiries that interfere with factual reconstruction. \\ \nopagebreak
    & Hate Speech or Incendiary Content & All points forfeited to zero & (0 Points) Explicitly discriminatory or stigmatizing inquiries (e.g., linking specific groups to criminal tendencies) or inducing the user to provide sensitive information. \\ \midrule

    \multirow{3}{\hsize}{\textbf{Truthfulness and Honesty:} The model should seek confirmation or pose clarifying questions when information is uncertain or contested.} 
    & Minor Imprecision & -1 to -2 Points & Contains 1-2 typos in the inquiry, or a date lacks the year (where only one year exists in the case facts), provided it does not impede comprehension. \\ \nopagebreak
    & Generalized Questioning (-3 points per instance, capped at -6 points) & -3 to -6 Points & (-3 Points) Repeatedly emphasizing “authenticity” or “whether you are concealing information” without pointing to specific contradictions or gaps; replacing effective inquiry with vague skepticism, leading to a failure to locate missing critical facts. \\ \nopagebreak
    & Inducing Fraud & All points forfeited to zero & (0 Points) Guiding the falsification of evidence or evasion of discovery (e.g., “You could backdate a contract or forge chat records”); or asserting facts directly when information is insufficient (treating speculation as fact). \\

\end{xltabular}

\clearpage
\twocolumn 
\justifying

\subsection{Rubrics of Practical Case Analysis Task}
\label{appendix:Details of Rubrics2}

The Practical Case Analysis task evaluates a model's capacity for complex legal reasoning and statutory application. For this task, experts curate authentic legal scenarios from judicial judgments and professional reports, constructing subjective questions that require a structured response. Each evaluation is grounded in a reference answer comprising four essential components: \textbf{Legal Conclusion, Fact Interpretation, Reasoning Process, and Statute Location}.

To ensure a high-degree of evaluative rigor, we have established a granular scoring system totaling approximately 11,000 rubric items, with an average of 14.1 items per task (See Details in \ref{tab:rubrics_task2}). These rubrics are meticulously categorized into \textbf{Sequential Reasoning} and \textbf{Non-sequential Reasoning} dimensions to reflect the inherent logic of legal practice. The formulation of these rubrics follows four core standards: 
\begin{itemize} 
    \item \textbf{Granularity}: Each item represents a single, discrete scoring point (e.g., identifying a party's legal standing based on specific ownership facts). 
    \item \textbf{Precision}: Scoring points must be legally accurate and formulated to eliminate interpretive ambiguity. 
    \item \textbf{Exclusivity}: Rubric items are non-redundant and collectively encompass the entire evaluative scope of the case. 
    \item \textbf{Quantification}: Points are allocated according to a predefined distribution, with the final score determined by the aggregate of validated rubric items. 
\end{itemize}

Furthermore, we define specific structural requirements for different reasoning patterns to assess the model's logical integrity (See Details in Table \ref{tab:rubrics_task2}): 
\begin{itemize} 
    \item \textbf{Non-Sequential Reasoning Requirements}: This applies when the order of argumentation does not affect the validity of the conclusion. For example, when determining whether a plaintiff's lawsuit meets the statutory filing conditions, the reasoning need only address the four elements stipulated in Article 122 of the Civil Procedure Law. There is no mandatory sequence for evaluating these elements. When drafting, do not label steps (e.g., “Step X”); instead, list them directly as “Analysis of [Element A],” “Analysis of [Element B],” etc. 
	\item \textbf{Sequential Reasoning Requirements}: This applies to issues that must follow a strict argumentative progression to maintain logical integrity. For instance, in copyright infringement cases involving images, the initial step must be to analyze whether the image in question constitutes a “protected work.” If the image does not qualify as a work, subsequent discussions regarding infringement become moot. Analyzing infringement directly without first establishing the status of the work fails to meet the sequential requirement. When drafting, explicitly state: “Step 1: Analysis of [Issue X]”; “Step 2: Analysis of [Issue Y].”  
\end{itemize}

Following the curation of these rubrics, the evaluation is executed through a standardized pipeline. First, a candidate model generates a legal response based on the case facts. Subsequently, a high-performance LLM acts as an automated judge, utilizing the expert-written reference answer as the ground truth. This evaluator model applies the granular scoring logic defined in the rubrics to objectively assess the candidate's performance across all reasoning dimensions.

\clearpage
\onecolumn 
\begin{sidewaystable} 
\centering
\small
\renewcommand{\arraystretch}{1.5} 

\newcolumntype{Y}{>{\hsize=0.6\hsize\RaggedRight\arraybackslash}X} 
\newcolumntype{S}{>{\hsize=0.7\hsize\RaggedRight\arraybackslash}X} 
\newcolumntype{Z}{>{\hsize=1.0\hsize\RaggedRight\arraybackslash}X} 
\newcolumntype{N}{>{\hsize=1.7\hsize\RaggedRight\arraybackslash}X} 

\begin{tabularx}{\textwidth}{Y | S | Z | N} 
    \noalign{\hrule height 0.8pt}
    \textbf{Dimension} & \textbf{Content} & \textbf{Point Allocation} & \textbf{Examples} \\ 
    \hline
    
        \textbf{Case Facts Scoring} & Evaluation of Facts Extraction & Identification of each critical case fact earns +5 points. Missing sub-elements within a key fact result in a 1–3 point deduction until the 5-point cap is reached. & [Conclusion Scoring] (+5) Pitaya Company has the right to demand that Banana Machinery Company pay the outstanding debt of 18.88 million RMB and the corresponding interest (failure to specify the amount of 18.88 million RMB results in a 2-point deduction; failure to mention interest results in a 2-point deduction). \\ \midrule

    \textbf{Reasoning Process}
    & Sequential Reasoning & Points are awarded only if the logic follows a specific, step-by-step order (10 Points per Step/Point). Evaluation is based purely on the logical progression of the argument, independent of the final conclusion. & [Reasoning Process Scoring] Scoring condition: Points are awarded only if reasoning follows the sequence of Step 1 and Step 2; no points are awarded if out of sequence. 

    Step 1: Analyze whether the involved images constitute works protected by the Copyright Law. (+10 points) 
    
        $\bullet$ Analysis of the requirement that the work belongs to the fields of literature, art, or science:  Mention that the images generated by Lin Yue through the use of prompts belong to creations in the field of art. 
        
        $\bullet$ Analysis of the requirement that the work be expressed in a certain form:  Mention that the images generated by Lin Yue through the use of prompts can be expressed in a certain form. 
        
        $\bullet$ Analysis of the requirement that the work constitutes an intellectual achievement:  Mention that the images generated by Lin Yue through the use of prompts are intellectual achievements created by a human. 
        
        $\bullet$ Analysis of whether the work possesses originality:  Mention Lin Yue's design of the prompts and the requirements for originality in existing judgments. 
        
    Step 2: Analyze whether the actions of Dongshan Trading Company, Shen Qing, and Hengtong Plastic Company infringe upon the copyright. (+10 points) 
    
        $\bullet$ Analysis of the ``Access'' element: Mention that the defendants' designs originated from independently generated drafts and third-party optimization; the prompts shared by the plaintiff belong to ``ideas'' rather than protected ``expressions,'' and the prompts are simple, lack distinctiveness, and contain elements previously presented. 
        
        $\bullet$ Analysis of the ``Substantial Similarity'' element: Mention that there are significant differences between the accused product and the plaintiff's images in specific levels of expression, such as wing curvature, patterns, chair leg shapes, and overall style; the defendants also conducted designs adapted for industrial production.
     \\

    \noalign{\hrule height 0.8pt}
\end{tabularx}
\caption{Rubrics of Task 2: Practical Case Analysis (I).}  \label{tab:rubrics_task2}
\label{tabapp:label_table_detailed_I}
\end{sidewaystable}
\clearpage

\clearpage
\begin{sidewaystable}
\centering
\footnotesize
\renewcommand{\arraystretch}{1.5} 

\newcolumntype{Y}{>{\hsize=0.4\hsize\RaggedRight\arraybackslash}X} 
\newcolumntype{S}{>{\hsize=0.5\hsize\RaggedRight\arraybackslash}X} 
\newcolumntype{Z}{>{\hsize=0.7\hsize\RaggedRight\arraybackslash}X} 
\newcolumntype{N}{>{\hsize=2.4\hsize\RaggedRight\arraybackslash}X} 

\begin{tabularx}{\textwidth}{Y | S | Z | N} 
    \noalign{\hrule height 0.8pt}
    \textbf{Dimension} & \textbf{Content} & \textbf{Point Allocation} & \textbf{Examples} \\ 
    \hline
    
    \textbf{Reasoning Process} & Non-Sequential Reasoning & Points are awarded for individual analytical points regardless of their order (10 Points per Step/Point). & [Reasoning Process Scoring] Analyze the four requirements according to Article 122 of the Civil Procedure Law:  
    
    Analysis regarding Plaintiff's Standing (+10 points) 

    $\bullet$  Mention that Party A is the owner of the involved Xiaomi mobile phone. 
    
    $\bullet$ Mention that Party B possesses the phone and refuses to return it, directly infringing upon Party A's ownership. 
    
    Analysis regarding the Specificity of the Defendant (+10 points) 
    
    $\bullet$ Mention that Party A's target of litigation is Party B, an individual industrial and commercial household providing screen replacement services. 
    
    $\bullet$ Mention that Party B's identity is definite, and the contact information and business address are clear (Party A was able to mail the phone to their repair station). 
    
    Analysis regarding Specific Litigation Requests, Facts, and Grounds (+10 points) 
    
    $\bullet$  Conduct an analysis of the specific litigation requests: Mention the litigation request”to order Party B to immediately return the involved Xiaomi mobile phone and bear the litigation costs of this case.” 
    
    $\bullet$  Conduct an analysis of the specific facts and grounds: Centering on the litigation requests, Party A stated a series of specific facts, including the establishment of a repair contract relationship between the two parties, the accidental mailing of the phone by themselves, and the detention of the phone by Party B without return. Party A claims that Party B's behavior is”without legal basis” and constitutes the legal grounds for the lawsuit. Simultaneously, Party A submitted preliminary evidence such as chat records and courier vouchers to support the claim. 
   
    Analysis regarding the Scope of Court Acceptance and Jurisdiction (+10 points) 
    
    $\bullet$ Mention that this case is a dispute arising from the possession and return of property, which is a property rights dispute and a typical civil dispute. 
    
    $\bullet$ Mention that according to the prompt information,”Party A filed a lawsuit in a court with jurisdiction,” which indicates that the issue of jurisdiction has been resolved and requires no further in-depth discussion. \\ \midrule

    \textbf{Statute Location} & Evaluation of Statute Location & Accurately citing the relevant law or specific article earns +5 points. & (+5 points) Article 23 of the Company Law of the PRC: Where a shareholder of a company abuses the independent status of the company as a legal person and the limited liability of shareholders to evade debts and seriously damages the interests of the creditors of the company, the shareholder shall bear joint and several liability for the debts of the company. Where a shareholder uses two or more companies under their control to commit the acts specified in the preceding paragraph, each company shall bear joint and several liability for the debts of any of the companies. In the case of a company with only one shareholder, if the shareholder cannot prove that the company's property is independent of the shareholder's own property, the shareholder shall bear joint and several liability for the debts of the company. \\

    \noalign{\hrule height 0.8pt}
\end{tabularx}
\caption{Rubrics of Task 2: Practical Case Analysis (II).}
\label{tabapp:label_table_detailed_I}
\end{sidewaystable}
\clearpage
\twocolumn 

\subsection{Rubrics of Legal Document Generation Task}
\label{appendix:Details of Rubrics3}

In rubric design, we reduce the weight assigned to formal presentation to less than 10\% and focus evaluation on substantive content. We emphasize models' ability to filter incorrect and irrelevant information and design case-specific rubrics for each scenario to ensure objective and effective evaluation of model outputs.

Specifically, the rubric of Plaintiff focuses on:

\begin{itemize}
    \item Procedures And Jurisdiction: Identify the Applicable Court, Determine Proper Party, Select the most Favorable Cause of Action
    \item Key Claims and Best Strategy of Claim: Ensuring Reasonableness and Completeness of the Prayer for Relief, Capability to Accurately Calculate Amounts
    \item The Construction of Facts by Evidence: Logical Structuring of Facts and Reasoning, Preliminary Organization of Evidence
\end{itemize}

The rubric of the Defendant focuses on:

\begin{itemize}
    \item Selection of Defense Strategies: Restructuring of Legal Relationships, Determining Core Defense Strategies, Demonstrates Solid and Coherent Argumentation
    \item Procedural Defenses of Jurisdiction and Admissibility: Jurisdictional Objections, Challenges to Legal Standing, Defenses Based on Procedural Timelines, Procedural Dismissal Motions
    \item Defenses Against Damages and Compensation Claims: Arguing of causation or intervening factors, Excluding non-compensable losses such as indirect damages, Challenging the calculation of plaintiff's claimed amount
\end{itemize}

In addition, to ensure compliance with the fundamental normative requirements of legal documents, a set of uniform penalty criteria has been designed and shall apply to all types of assessed documents, such as statements of claim and defense. The uniform penalty criteria follow a cumulative deduction rule, with deductions accumulating up to the total score of the question. The specific criteria are as follows:

\begin{itemize}
    \item Accuracy of Provisions Citation: The serial numbers and content of legal provisions cited in the document must be authentic, valid, and accurate. Each error in a legal provision (e.g., incorrect serial number, misrepresentation of content, citation of an invalid provision, etc.) will result in a 10-point deduction.
    \item Compliance with Instructions: The document must be generated strictly in accordance with user instructions or the template of the specified document type. If the document type generated does not match the required type (e.g., a statement of defense is requested but a statement of claim is produced), a 10-point deduction will apply.
    \item Document Standards: The following standards must be met simultaneously. For each standard not satisfied, a 10-point deduction will apply:
    \begin{itemize}
        \item The format complies with statutory format requirements for the corresponding type of legal document.
        \item Wording is precise and standardized, using formal legal terminology without idiom or ambiguous expressions.
        \item All essential elements of the document are included, with no omission of core components (e.g., absence of party information, or lack of a clear statement of defense).
    \end{itemize}
\end{itemize}

\clearpage
\onecolumn 

\begin{sidewaystable}
\centering
\small
\renewcommand{\arraystretch}{1.5}

\newcolumntype{Y}{>{\hsize=0.4\hsize\RaggedRight\arraybackslash}X} 
\newcolumntype{S}{>{\hsize=0.8\hsize\RaggedRight\arraybackslash}X} 
\newcolumntype{Z}{>{\hsize=1.8\hsize\RaggedRight\arraybackslash}X} 

\begin{tabularx}{\textwidth}{Y | S | Z } 
    \noalign{\hrule height 0.8pt}
    \textbf{Dimension} & \textbf{Content} & \textbf{ID/Point}  \\ 
    \hline

    \rowcolor[gray]{0.95} \multicolumn{3}{c}{\textbf{The Rubric of Plaintiff}} \\ \midrule
    \multirow{3}{0.8\hsize}{\textbf{Procedures And Jurisdiction}} 
    & $\bullet$ Identify the Applicable Court & Following with the Civil Procedure Law and relevant rules, accurately determine the court with territorial and hierarchical jurisdiction to ensure the litigation proceedings.	 \\* \nopagebreak
    & $\bullet$ Determine Proper Party & Clearly identify the plaintiff's and defendant's legal interest, ensure the legitimacy of the parties' standing, and guarantee the accuracy and completeness of their information.	 \\* \nopagebreak
    & $\bullet$ Select the most Favorable Cause of Action & Precisely define the legal relationships, select suitable cause of action, and prioritize the cause of action most favorable to one's own side. \\ \midrule

    \multirow{2}{0.8\hsize}{\textbf{Key Claims and Best Strategy of Claim}} 
    & $\bullet$ Ensuring Reasonableness and Completeness of the Prayer for Relief & Translate the client's interests into specific and enforceable claims, comprehensively covering core rights and ancillary rights.	 \\* \nopagebreak
    & $\bullet$ Capability to Accurately Calculate Amounts & Accurately calculate the amounts related to the claims, ensuring that the calculation basis, standards, and commencement and cutoff dates are lawful, and that the process is clear and verifiable.	 \\ \midrule

    \multirow{2}{0.8\hsize}{\textbf{The Construction of Facts by Evidence}} 
    & $\bullet$ Logical Structuring of Facts and Reasoning & Narrate the case facts clearly along a timeline, highlighting key events, and achieve precise alignment between facts and legal grounds.	 \\* \nopagebreak
    & $\bullet$ Preliminary Organization of Evidence & Organize the evidence list, clarify the purpose of proof, and ensure that each piece of evidence corresponds to the factual statements, without omitting any core or key evidence.	 \\ \midrule

    \rowcolor[gray]{0.95} \multicolumn{3}{c}{\textbf{The Rubric of Defendant}} \\ \midrule
    \multirow{3}{0.8\hsize}{\textbf{Selection of Defense Strategies}} 
    & $\bullet$ Restructuring of Legal Relationships & Break through the legal relationship framework established by the plaintiff's claims, reconstruct a legal relationship favorable to one's own side.	 \\* \nopagebreak
    & $\bullet$ Determine Core Defense Strategies & Identify key winning arguments in the case. For example: arguing for the invalidity of the contract, claiming force majeure or fundamental change of circumstances, or invoking the defense of prior performance.	 \\* \nopagebreak
    & $\bullet$ Demonstrates Solid and Coherent Argumentation & Ensure that all defense arguments are consistent and free of contradictions, and that a complete logical chain from “facts → law → conclusion” is formed, thereby guaranteeing the coherent argumentation.	 \\ \midrule

    \noalign{\hrule height 0.8pt}
\end{tabularx}
\caption{Rubrics of Task 3: Legal Document Generation (I).}
\label{tabapp:label_table_detailed_I} 
\end{sidewaystable}
\clearpage
\twocolumn 

\clearpage
\onecolumn 

\begin{sidewaystable}
\centering
\small
\renewcommand{\arraystretch}{1.5}

\newcolumntype{Y}{>{\hsize=0.4\hsize\RaggedRight\arraybackslash}X} 
\newcolumntype{S}{>{\hsize=0.8\hsize\RaggedRight\arraybackslash}X} 
\newcolumntype{Z}{>{\hsize=1.8\hsize\RaggedRight\arraybackslash}X} 

\begin{tabularx}{\textwidth}{Y | S | Z } 
    \noalign{\hrule height 0.8pt}
    \textbf{Dimension} & \textbf{Content} & \textbf{ID/Point}  \\ 
    \hline

\multirow{4}{0.8\hsize}{\textbf{Procedural and Jurisdictional Defenses}} 
    & $\bullet$ Jurisdictional Objections & Identify jurisdiction errors or exclude court jurisdiction based on arbitration agreements; accurately distinguish the jurisdictional boundaries of criminal-involved cases to secure procedural advantages.	 \\* \nopagebreak
    & $\bullet$ Challenges to Legal Standing & Examine and challenge the legitimacy of the plaintiff's standing to sue and the propriety of the defendant as a party.	 \\* \nopagebreak
    & $\bullet$ Defenses Based on Procedural Timelines & Identify defense opportunities based on key time points such as the statute of limitations and exclusion periods.	 \\* \nopagebreak
    & $\bullet$ Procedural Dismissal Motions & For example, apply for a suspension of proceedings (such as in cases where criminal proceedings take precedence over civil ones, or while awaiting the outcome of another case), or argue that the plaintiff has not exhausted prerequisite remedies (such as the confirmation procedure in state compensation claims).	 \\ \midrule

    \multirow{3}{0.8\hsize}{\textbf{Defenses Against Damages and Compensation Claims}} 
    & $\bullet$ Arguing of causation or intervening factors & Through evidence or argumentation, sever the direct causal link between one's own actions and the alleged harm, or assert that intervening factors caused the harm.	 \\* \nopagebreak
    & $\bullet$ Excluding non-compensable losses such as indirect damages & In accordance with legal provisions, exclude non-compensable losses such as indirect losses and lost profits, and confine the scope of compensation to reasonable limits.	 \\* \nopagebreak
    & $\bullet$ Challenging the calculation of plaintiff's claimed amount & Challenge the calculation basis, standards, or other elements of the amount claimed by the plaintiff, assert applicable rules such as set-off of gains and losses, and seek to reduce the compensation amount.	 \\ \midrule

    \rowcolor[gray]{0.95} \multicolumn{3}{c}{\textbf{The Rubric of Penalty}} \\ \midrule
    \textbf{Penalty Item} & \multicolumn{2}{l}{\textbf{Specific Requirements}} \\ \midrule
    Accuracy & Accurately citing the specific article & The serial numbers and content of legal provisions cited in the document must be authentic, valid, and accurate. Each error in a legal provision (e.g., incorrect serial number, misrepresentation of content, citation of an invalid provision, etc.) will result in a 10-point deduction.		 \\ \midrule
    Compliance with Instructions & Following formatting and submission rules & The document must be generated strictly in accordance with user instructions or the template of the specified document type. If the document type generated does not match the required type (e.g., a statement of defense is requested but a statement of claim is produced), a 10-point deduction will apply.		 \\ \midrule
    \multirow{3}{0.8\hsize}{Document Standards} 
    & $\bullet$ Professional tone and language & The format complies with statutory format requirements for the corresponding type of legal document.		 \\* \nopagebreak
    & $\bullet$ Logical document structure & Wording is precise and standardized, using formal legal terminology without idiom or ambiguous expressions.		 \\* \nopagebreak
    & $\bullet$ Proper legal formatting & All essential elements of the document are included, with no omission of core components (e.g., absence of party information, or lack of a clear statement of defense).		 \\

    \noalign{\hrule height 0.8pt}
\end{tabularx}
\caption{Rubrics of Task 3: Legal Document Generation (II).}
\label{tabapp:label_table_detailed_I}
\end{sidewaystable}
\clearpage
\twocolumn 

\section{Model Details}
\label{appendix:model_details}

In this section, we provide brief introductions to the 10 large language models evaluated in our experiments.

\begin{description}
    \item[Claude-Sonnet-4.5]~\cite{anthropic2025claude45} The latest iteration in Anthropic's Claude series, Sonnet-4.5 balances high performance with cost-efficiency, demonstrating enhanced capabilities in nuanced instruction following and complex reasoning tasks.
    
    \item[Gemini-2.5-Pro]~\cite{comanici2025gemini} An upgraded version of Google's multimodal model, Gemini-2.5-Pro features improved long-context understanding and more robust performance across diverse logic and coding benchmarks.
    
    \item[Gemini-3.0-Pro-Preview]~\cite{pichai2025gemini3} The next-generation preview model from the Gemini family, designed with advanced native multimodal integration and superior reasoning depth for complex agentic tasks.
    
    \item[GPT-4o-20240806]~\cite{katz2024gpt} A highly optimized snapshot of OpenAI's GPT-4o, known for its rapid inference speed and state-of-the-art performance in multimodal processing and structured output generation.
    
    \item[GPT-5-0807-Global] The foundational version of the GPT-5 series, representing a significant leap in general intelligence, with substantial improvements in world knowledge and reasoning stability compared to its predecessors.
    
    \item[GPT-5.2-1211-Global] An iterated and refined version of the GPT-5 architecture, offering enhanced instruction adherence and reduced hallucination rates, specifically tuned for complex problem-solving.
    
    \item[DeepSeek-V3.2]~\cite{liu2025deepseek} Developed by DeepSeek, this model utilizes a Mixture-of-Experts (MoE) architecture to achieve competitive performance with top-tier proprietary models while maintaining high inference efficiency.
    
    \item[Kimi-K2]~\cite{team2025kimi} The second-generation model from Moonshot AI, Kimi-K2 specializes in lossless long-context processing, capable of handling massive information retrieval and synthesis tasks effectively.
    
    \item[Qwen3-235b-Think]~\cite{yang2025qwen3} The latest version of the Qwen model family. We use qwen3-235b-a22b-think, a flagship model featuring a unified framework that dynamically allocates a “thinking budget” to balance latency and performance, specifically optimized for complex multi-step reasoning.
    
    \item[Qwen3-Max] The most capable model in Alibaba's Qwen3 lineup, designed to handle the most demanding tasks with comprehensive knowledge coverage and superior complex agent planning abilities.
\end{description}

\section{Prompt Engineering}
\label{sec:app_prompts}

We utilized specific prompts to guide the models in generating responses and to instruct the Judge model for evaluation. Tables \ref{tab:prompt_task1}, \ref{tab:prompt_task2}, \ref{tab:prompt_task3},through \ref{tab:score_prompt_task1}, \ref{tab:score_prompt_task2}, \ref{tab:score_prompt_task3} display the core content of these prompts.

\subsection{Response Generation Prompts}

\subsubsection{Prompt Design for Task 1: Public Legal Consultation}

The prompt for Task 1 transitions the LLM from a passive information provider to an active legal interrogator by enforcing a rigorous fact-finding protocol (As shown in Tables \ref{tab:prompt_task1}). This design requires the model to reconstruct the full scope of an event by prioritizing the Seven Elements of Evidence, which include time, location, parties, actions, amounts, evidence, and sources. To address the inherent bias in user narratives, the prompt incorporates a Concealment Detection mechanism that utilizes adversarial and reciprocal questioning to probe for latent contradictions or unfavorable facts. Finally, the model is mandated to maintain a logical hierarchy by addressing high-stakes issues such as personal safety and property title before supplementary details. This structured approach ensures professional neutrality and provides the granularity necessary for complex litigation strategy.

This is an example of adapted case (Shown in Table \ref{tab:adapted_case}). The adapted case incorporates subtle factual distortions and selective disclosures to obfuscate the legal nature of the relationship between the parties. By introducing personal affiliations (the romantic relationship), emphasizing formalistic but non-determinative evidence (access to the corporate seal and third-party emails), and detailing a circular financial arrangement for social insurance payments, the narrative creates a facade of a legitimate labor relationship. These "traps" are designed to test the analyst's ability to look beyond surface-level administrative records and identify the absence of substantive labor elements, such as organizational subordination, actual wage payment by the employer, and the existence of a bona fide job position.

\subsubsection{Prompt Design for Task 2: Practical Case Analysis}

The prompt for Task 2 evaluates the ability of Large Language Models to perform systematic legal deconstruction by enforcing a structured cognitive framework aligned with professional judicial standards (As shown in Tables \ref{tab:prompt_task2}). This design mandates a strict “Conclusion First” response order, followed by factual extraction, logical reasoning, and statutory citations to mirror the organizational style of professional legal memorandums. This structured reasoning chain creates multiple points of verification, ensuring granular transparency and allowing practitioners to identify specific points where a model might diverge from established judicial logic. Furthermore, the step-by-step requirement necessitates a direct alignment between case facts and specific legal provisions, providing an explicit mapping that allows experts to audit the precision of the reasoning. This mechanism effectively addresses the reasoning gap by preventing models from reaching correct conclusions through flawed or hallucinated legal paths.

\subsubsection{Prompt Design for Task 3: Legal Document Generation}

The prompt for Task 3 evaluates the proficiency of Large Language Models in professional legal drafting by requiring the transformation of informal client narratives into formal legal instruments (As shown in Tables \ref{tab:prompt_task3}). A primary feature of this design is the requirement for autonomous document classification, where the model must independently determine the appropriate document type, such as a Bill of Complaint or a Statement of Defense, based on the procedural roles identified within the context. This requires the model to distinguish between the grievances of a plaintiff and the rebuttals of a defendant before applying the corresponding legal template. Furthermore, the prompt mandates a critical analysis of user input to identify and correct factual inconsistencies or potential traps resulting from low legal literacy. By filtering out irrelevant emotional descriptions while retaining legally significant facts, the model ensures that the final document remains legally sound and compliant with the statutory standards and judicial norms of Mainland China.

\clearpage
\onecolumn
\begin{table*}[h]
\centering
\small
\begin{tabularx}{\linewidth}{>{\raggedright\arraybackslash}X}
\toprule
\textbf{Task 1 Prompt: Public Legal Consultation} \\
\midrule
\textbf{\textsc{Role}} \\
You are a senior attorney with over a decade of practice experience, possessing expert-level knowledge of current Chinese laws, regulations, and judicial practices. You excel at reconstructing the full scope of an event based on the one-sided facts provided by a client, identifying potential concealments or biases, and completing the evidentiary chain and timeline to prepare for subsequent legal analysis and litigation strategy. \\
\midrule
\textbf{\textsc{Core Requirements}} \\
1. \textbf{Neutrality:} Remain objective and non-suggestive. Avoid taking sides or using qualitative/judgmental labels (e.g., ``scumbag,'' ``abuser,'' or ``certainly''). \\
2. \textbf{Verifiability and Evidence-Orientation:} Focus inquiries on ``Time, Location, Parties, Actions, Amounts, Evidence, and Sources.'' \\
3. \textbf{Comprehensive Coverage:} Cover all aspects of the event, details of signed agreements, mutual faults, and potential risk points. \\
4. \textbf{Prioritize Critical Issues:} Address questions that most significantly impact the case outcome first (e.g., safety, evidence, property ownership, nature of debt, or custody arrangements) before moving to supplementary details. \\
5. \textbf{Multi-Question Output:} Provide at least 25 questions per response in a numbered list. Questions must be concise, clear, and focused on a single point of inquiry. \\
6. \textbf{Identify Contradictions or Gaps:} You must include clarification questions to address latent contradictions (e.g., inconsistent timelines, unclear sources of funds, or questionable authenticity of agreements). \\
7. \textbf{Concealment Detection:} Proactively inquire about potentially omitted unfavorable facts using ``reciprocal questioning'' (e.g., ``Was it a mutual physical altercation?'', ``Who initiated the physical contact?'', ``Are there police records?'', ``Was there an extramarital affair?'', ``Was the loan jointly signed?''). \\
8. \textbf{Output Format:} Provide only the list of questions. Do not include any introductory or concluding text. \\
\midrule
\textbf{\textsc{Response Example}} \\
1. Does [Name] hold a Special Operation Certificate for welding? \\
2. In the agreement regarding ``[Subject]'' between [Party A] and [Party B], was the relationship defined as employment or a dynamic contract for work? \\
3. After the 2,000 RMB was transferred to [Name]'s personal WeChat account, was there a written agreement stating this was ``collection on behalf of the company'' or any subsequent vouchers for company receipts? \\
4. Does the outer packaging of the tea involved in this case label the entrusting party as [Company X] and the entrusted party as [Factory Y]? \\
5. ...... \\
6. ...... \\
7. ...... \\
\midrule
\textbf{\textsc{User Input Format}} \\
``You are an attorney. Your client is consulting with you and states: \texttt{<conversation>}. You are now required to raise follow-up questions to your client to determine the full scope of the event.'' \\
\bottomrule
\end{tabularx}
\caption{System prompt for Legal Consultation task.}
\label{tab:prompt_task1}
\end{table*}
\clearpage
\twocolumn

\clearpage

\newcolumntype{J}{>{\justifying\arraybackslash}X}

\onecolumn
\renewcommand{\arraystretch}{1.5} 
\begin{xltabular}{\textwidth}{J | J}
\caption{Comparison of Case Details} \label{tab:adapted_case} \\

\toprule
\rowcolor{gray!15} 
\textbf{Raw User’s Narrative Version} & \textbf{Bar Exam Version} \\ \midrule
\endfirsthead

\multicolumn{2}{c}{continued Table \ref{tab:adapted_case}} \\
\toprule
\rowcolor{gray!15} 
\textbf{Raw User’s Narrative Version} & \textbf{Bar Exam Version} \\ \midrule
\endhead

\midrule
\multicolumn{2}{r}{\textit{Continued on next page...}} \\
\footnotemark
\endfoot

\bottomrule
\endlastfoot

It all started back in November 2017. At that time, I was introduced to Hengyuan International Trade Co., Ltd. by a friend named Lin Tao, who had grown up with me since childhood (Hidden case detail: Lin Tao and the party involved were in a romantic relationship; this detail has no impact on the case determination, but adding this information makes it easier to identify the fabricated labor relationship). He and I were extremely close—our parents were old neighbors—and he was already a shareholder in the company at that time, holding 30\% of the shares, so I trusted him completely. He told me that the company had just started and urgently needed a reliable person to help the general manager handle core affairs, so he had me take the position of general manager’s assistant. \par \vspace{0.5em}

On my first day, the general manager personally received me and talked with me in his office for nearly an hour. He explicitly said: ‘Your salary will be 16,000 yuan per month. We’ll settle on it verbally for now, and the contract will be supplemented next week.’ (No such position actually existed; this was a statement colluded between Lin and the manager. Adding this case detail makes it easier to identify the fabricated labor relationship.) I also saw the company’s business license and official seal on his desk, which made me feel that everything was very formal and legitimate. From that day on, I devoted myself wholeheartedly to the work, going out early and returning late every day, helping the general manager organize customer data, draft meeting minutes, and coordinate with three key foreign trade customers—these customers even sent me thank-you emails later! (The so-called ‘coordinating with customers’ was actually business from another company controlled personally by Lin Tao, unrelated to Hengyuan Trade, but the case description deliberately blurred this as ‘company customers.’ Adding this case detail makes it easier to identify the fabricated labor relationship!) & 

In November 2017, Qiao was introduced by her boyfriend Lin Tao (who was also a shareholder of Hengyuan International Trade Co., Ltd. [hereinafter “Hengyuan Company”], holding 30\% of shares) to join Hengyuan Company as a general manager’s assistant. During her employment, Qiao was responsible for organizing customer data, drafting meeting minutes, and coordinating with three foreign trade customers (which were actually business operations controlled personally by Lin Tao). However, Hengyuan Company did not implement attendance tracking for Qiao, did not issue her an employee badge or system account, and did not sign a labor contract with her. \\ \midrule

But who could have imagined that not a single penny of salary was ever paid? At first, Lin Tao comforted me, saying: ‘The company’s cash flow is tight, but I will definitely get your social insurance set up first—don’t worry.’ (The social insurance fees were actually transferred in cash from Qiao Xiaoli to Lin Tao, who then paid them on behalf of the company in the company’s name; the company did not actually bear the costs. Her pregnancy at the time of employment should be noted for disclosure.) Sure enough, from November 2017 to June 2018, my social insurance was normally paid. I checked the system records myself—the insured unit clearly stated ‘Hengyuan International Trade Co., Ltd.’ Doesn’t this prove that I was their employee? \par \vspace{0.5em}

In early 2018, I found out I was pregnant and immediately told the general manager. He even congratulated me, saying, ‘The company supports you—take your maternity leave according to regulations, and your salary will not be reduced by a single cent.’ I started my maternity leave at the end of April and gave birth to my baby smoothly in May (I was already pregnant at the time of employment, in order to qualify for maternity subsidies and protection). But up to now, not a single penny of maternity leave salary has been received. When I asked the finance department, they stammered and said, ‘We haven’t received any salary payment instructions.’ (No maternity certificates, maternity leave application forms, or other required documents were submitted to the company.) \par \vspace{0.5em}

[Omission due to length] & 

From November 2017 to June 2018, at Qiao’s request and with her bearing the social insurance costs herself, Hengyuan Company normally paid social insurance for Qiao, with the social insurance system showing Hengyuan Company as the participating unit. In early 2018, Qiao informed the general manager about her pregnancy, and she began maternity leave at the end of April. After giving birth in May, Hengyuan Company did not pay her maternity leave wages. On July 12, 2018, the general manager notified Qiao via WeChat to terminate the labor relationship on the grounds of “optimizing personnel structure,” without providing a written termination notice. Thereafter, Hengyuan Company stopped paying Qiao’s social insurance and refused to process social insurance reduction procedures, causing Qiao to miss two job opportunities. Qiao filed claims requesting the company to pay wages, double salary difference for unsigned contract, and other amounts totaling 290,000 yuan, and to issue a certificate of termination. \\ \midrule

\multicolumn{2}{J}{\textbf{Question:} Based on the above case details, analyze whether each of Qiao's litigation claims can be supported, and explain the reasons.} \\

\end{xltabular}

\clearpage
\twocolumn

\clearpage
\onecolumn
\begin{xltabular}{\linewidth}{>{\RaggedRight\arraybackslash}X}
    \caption{System prompt for Case Analysis task.} \label{tab:prompt_task2} \\
    
    \toprule
    \textbf{Task 2 Prompt: Practical Case Analysis} \\
    \midrule
    \endfirsthead

    \multicolumn{1}{c}{{\bfseries \tablename\ \thetable{} -- Continued from previous page}} \\
    \toprule
    \textbf{Task 2 Prompt: Practical Case Analysis (Continued)} \\
    \midrule
    \endhead

    \midrule
    \multicolumn{1}{r}{{Continued on next page...}} \\
    \endfoot

    \bottomrule
    \endlastfoot

    \textbf{\textsc{Role}} \\
    You are a legal practice expert with over a decade of experience, possessing mastery over current Chinese laws, regulations, and judicial practices. You specialize in deconstructing complex legal issues into clear logical modules and providing answers strictly adhering to a professional style characterized by ``Conclusion First, Facts as the Foundation, Rigorous Reasoning, and Statutory Support.'' \\
    \midrule
    \textbf{\textsc{Core Requirements}} \\
    1. \textbf{Strict Sequence:} The response must unfold in the following order using the corresponding headings: \texttt{[Conclusion]}, \texttt{[Case Facts]}, \texttt{[Reasoning Process]}, \texttt{[Statutory Basis]}. \\
    2. \textbf{Content Standards:} \\
    \quad $\bullet$ \textbf{Conclusion:} Provide a direct and clear affirmative or negative judgment regarding the core dispute of the inquiry. \\
    \quad $\bullet$ \textbf{Case Facts:} Based on the user's account, extract the factual elements relevant to the legal judgment concisely and objectively. Do not add assumptions or conjectures. \\
    \quad $\bullet$ \textbf{Reasoning Process:} Demonstrate step-by-step how the facts link to the statutes to derive the conclusion. Use a layered or bulleted format. \\
    \quad $\bullet$ \textbf{Statutory Basis:} Cite current and effective laws, judicial interpretations, or administrative regulations of Mainland China. Specify the issuing authority, title, and specific article, paragraph, or item. Provide the original text of the statute. \\
    \midrule
    \textbf{\textsc{Template}} \\
    \texttt{[Conclusion]} \\
    {[Provide a clear and definite legal judgment here]} \\
    \texttt{[Case Facts]} \\
    {(Extract only key facts relevant to the conclusion; remain objective and concise)} \\
    \texttt{[Reasoning Process]} \\
    1. ......; \\
    2. ......; \\
    \texttt{[Statutory Basis]} \\
    Article X, Paragraph X of the XXX Law (Issuing Authority: XXX) \\
    (If multiple bases exist, arrange them by hierarchy of authority or relevance) \\
    \midrule
    \textbf{\textsc{Response Example}} \\
    \texttt{[Conclusion]} \\
    1. The matrimonial property relationship shall be governed by the laws of the Macau SAR... \\
    2. Property I (Mainland City A) is identified as joint matrimonial property. \\
    3. Property II (Mainland City B) is identified as the personal property of Party R... \\
    \vspace{0.5em}
    \texttt{[Case Facts]} \\
    $\bullet$ \textbf{Key Fact 1:} Party L (Macau resident) and Party R (Mainland resident) registered marriage in Macau (2013) and divorced (2016)... \\
    $\bullet$ \textbf{Key Fact 2:} Signed ``Prenuptial Agreement Record'' stipulating Macau as habitual residence and ``General Community of Property''... \\
    $\bullet$ \textbf{Key Fact 3:} Property I purchased by Party L before marriage, registered solely to Party L... \\
    $\bullet$ \textbf{Key Fact 4:} Property II purchased by Party R's parents via housing reform; Party L consented to registration under Party R... \\
    \vspace{0.5em}
    \texttt{[Reasoning Process]} \\
    \textbf{Step 1: Determine applicable law.} According to Art. 24 of the \textit{Law of the Application of Law for Foreign-related Civil Relations}: parties may choose the law of habitual residence. Parties chose Macau; thus, Macau law applies. \\
    \textbf{Step 2: Determine property regime.} Macau Civil Code Art. 1609: community property consists of all present/future property unless excluded by law. \\
    \textbf{Step 3: Ownership of Property I.} Under ``General Community,'' pre-marital purchase (Property I) is not excluded; thus, it is joint property. \\
    \textbf{Step 4: Ownership of Property II.} Involves seniority welfare. Party L's waiver indicates recognition of title. Per Macau Civil Code Art. 1610(1)(a), property gifted with exclusion clauses is personal. Property II is Party R's personal property. \\
    \vspace{0.5em}
    \texttt{[Statutory Basis]} \\
    $\bullet$ Article 24 of the \textit{Law of the Application of Law for Foreign-related Civil Relations of the PRC}... \\
    $\bullet$ Article 1609 of the \textit{Civil Code of Macau}... \\
    $\bullet$ Article 1610 of the \textit{Civil Code of Macau}... \\
    \midrule
    \textbf{\textsc{User Input Format}} \\
    \texttt{<questions>} Please provide your answer in the following format: \texttt{[Conclusion]}+\texttt{[Case Facts]}+\texttt{[Reasoning Process]}+\texttt{[Statutory Basis]} \texttt{<asking>} \\
\bottomrule
\end{xltabular}
\clearpage
\twocolumn

\clearpage
\onecolumn
\begin{table*}[h]
\centering
\small 
\renewcommand{\arraystretch}{1.3} 
\begin{tabularx}{\linewidth}{>{\raggedright\arraybackslash}X}
\toprule
\textbf{Task 3 Prompt: Legal Document Generation} \\
\midrule
\textbf{\textsc{Role}} \\
You are a legal practice expert familiar with the formal drafting standards for legal instruments. Based on the statements or information provided by your client (party involved), you are required to generate a legal document. \\
\midrule
\textbf{\textsc{1. Document Types}} \\
Please generate the appropriate legal document based on the following options (ensure the content meets the specific requirements of the selected type): \\
\quad (1) Application (Bill of Complaint) \\
\quad (2) Defendant (Statement of Defense) \\
\midrule
\textbf{\textsc{2. Document Requirements}} \\
1. \textbf{Template Adherence:} Drafting must automatically follow the specific template provided for each case. If no specific template exists for a particular cause of action, use the general standard legal format used in Mainland China. \\
2. \textbf{Length Constraint:} The generated legal document must be between 2,500 and 3,000 characters in length. \\
3. \textbf{Concise Output:} Output the document text only. Do not include any “fluff,” such as conversational filler or simulated lawyer-client dialogue (e.g., “Certainly, here is your document...”). \\
4. \textbf{Autonomous Selection:} You must independently determine which of the three document types is appropriate based on the provided prompt and context. \\
5. \textbf{Critical Analysis:} The client may have low legal literacy, and their statements may not be entirely accurate or may contain errors. You must identify potential ``traps'' or factual inconsistencies and draft a document that is legally sound, reasonable, and compliant. \\
6. \textbf{Statutory Citations:} You must cite authentic and currently effective laws and regulations in the document. \\
\midrule
\textbf{\textsc{User Input Format}} \\
``You are an attorney. Your client has approached you and described the situation as follows: \texttt{<conversation>}. Based on the current scenario, please draft a Bill of Complaint/Statement of Defense for your client.'' \\
\bottomrule
\end{tabularx}
\caption{System prompt for Document Generation task.}
\label{tab:prompt_task3}
\end{table*}
\clearpage
\twocolumn

\subsection{Judge Model Scoring Prompts}
\label{sec:app_scoring_prompts}

\subsubsection{Scoring Prompt 1: Public Legal Consultation}

To provide an objective and scalable assessment of model performance, we developed a standardized scoring protocol for Task 1 that transitions the evaluator into a neutral judge (As shown is \ref{tab:score_prompt_task1}). This process is grounded in a strict literal comparison between the generated questions and expert-curated rubrics, ensuring that the final score directly reflects the presence of specific legal inquiry points rather than subjective interpretations. The mechanism employs a nuanced matching logic where full credit is awarded for complete semantic consistency, while partial credit is granted for responses that capture core themes but lack specific qualifiers. Furthermore, the protocol enforces strict neutrality and mandates a structured analysis for every scoring item.

\subsubsection{Scoring Prompt 2: Practical Case Analysis}

The evaluation protocol for Task 2 employs a rigorous item-by-item inspection mechanism to assess the quality of complex legal case analyses (As shown is \ref{tab:score_prompt_task2}). This system requires the evaluator to segment model responses into four distinct functional components comprising the conclusion, statutory basis, case facts, and reasoning process. By applying specialized rubrics to each section independently, the judge adheres to a strict literal matching principle that awards points only for explicitly mentioned information while prohibiting subjective inferences or supplementary reasoning on behalf of the model. The scoring workflow follows a structured deconstruction of the legal syllogism, ensuring that the major premise, minor premise, and logical application are each validated against expert criteria. This granular approach is finalized through a standardized JSON output format that details the breakdown of points, specific textual mentions, and analytical rationales.

\subsubsection{Scoring Prompt 3: Legal Document Generation}

The scoring protocol for Task 3 evaluates the proficiency of Large Language Models in generating formal legal instruments that adhere to professional drafting standards (As shown is \ref{tab:score_prompt_task3}). This evaluation process assesses the model's ability to autonomously select the correct document type, such as a Bill of Complaint or Statement of Defense, based on the procedural context of the user narrative. The judge measures the output against stringent formal requirements, including strict adherence to Mainland China legal templates, authentic statutory citations, and a specific length constraint of 2,500 to 3,000 characters. Furthermore, the scoring mechanism accounts for the model's capacity for critical analysis, specifically its ability to identify and rectify factual inconsistencies or legal traps within informal client statements. By comparing the generated document against expert-curated rubrics, the evaluator ensures that the final output is not only concise and devoid of conversational filler but also legally sound and compliant with judicial norms.

\clearpage
\onecolumn

\begin{table*}[h]
\centering
\small
\begin{tabularx}{\linewidth}{>{\raggedright\arraybackslash}X}
\toprule
\textbf{Scoring Prompt 1: Public Legal Consultation} \\
\midrule
\textbf{\textsc{Role}} \\
You are an evaluator. Your task is to score a model's output by comparing it against a standard answer containing specific scoring items to assess the model's performance. \\
\midrule
\textbf{\textsc{Input and Output Formats}} \\
Input consists of two parts: \\
1. Agent's Question List: The list to be evaluated (10–25 numbered items). \\
2. Standard Rubric: (Total Score: N points; each point is formatted as ``(+X points) Point Description''). \\
Output must follow this fixed format: \\
The response score is XX points. Total score is XX, with a scoring rate of XX\%. \\
The analysis is as follows: \\
$\bullet$ Point 1: [Scoring Status], Rationale: ....... \\
$\bullet$ Point 2: [Scoring Status], Rationale: ...... \\
$\bullet$ ...... \\
\midrule
\textbf{\textsc{Scoring Rules}} \\
1. \textbf{Literal Comparison:} You do not need to judge the accuracy of the answer itself. Your sole task is to strictly follow the standard answer, determine whether the model's response contains identical or substantively identical content, and calculate the final score. \\
2. \textbf{Strict Neutrality:} Do not suggest new questions, provide legal advice, evaluate the parties involved, or restate lengthy background information. Provide only the comparison and scoring explanations. \\
3. \textbf{Matching Logic:} For each “Question Point” in the standard rubric, search for the best match within the agent's question list. Award points based on the degree of matching: \\
\quad o 100\% Score: If the response fully covers key elements and is semantically consistent. \\
\quad o Partial Credit: If the response hits the core theme but lacks key qualifiers. For example, if the standard requires “police records, hospitalization records, and medical certificates” but the response only asks about “police records,” award partial credit at your discretion. \\
\quad o 0\% Score: If the response fails to cover the semantic key points. \\
\textbf{Summary:} Aggregate the scores and calculate the Scoring Rate = Earned Points / Total Points. \\
\midrule
\textbf{\textsc{User Input}} \\
The consultation scenario between the attorney and the client is \texttt{<conversation>}. In this scenario, the scoring rubric for the attorney's performance is \texttt{<rubrics>}. Please evaluate the following questions posed by the attorney: \texttt{<answers>}. \\
\bottomrule
\end{tabularx}
\caption{Judge prompt for scoring Task 1: Public Legal Consultation.}
\label{tab:score_prompt_task1}
\end{table*}
\clearpage
\twocolumn

\clearpage
\onecolumn
\begin{table*}[h]
\centering
\small
\renewcommand{\arraystretch}{1.25}
\begin{tabularx}{\linewidth}{>{\raggedright\arraybackslash}X}
\toprule
\textbf{Scoring Prompt 2: Practical Case Analysis} \\
\midrule
\textbf{\textsc{Role}} \\
You are a rigorous legal practice evaluation expert, specializing in item-by-item inspection of legal practice responses based on precise grading criteria. You strictly adhere to the principle of ``Award points if mentioned; deduct points if omitted,'' and you must avoid making subjective inferences. \\
\midrule
\textbf{\textsc{Core Task}} \\
Evaluate legal practice responses accurately according to the provided detailed scoring rubrics. A response is divided into four sections: Conclusion, Statutory Basis, Factual Summary, and Reasoning Process. Each section may have its own independent rubric. \\
\midrule
\textbf{\textsc{Input Information Structure}} \\
\textbf{Question Content:} A brief case description and the specific inquiry. \\
\textbf{Scoring Rubric:} Divided into multiple parts. The format for each may follow this example: \\
$\bullet$ Conclusion Rubric \\
$\bullet$ Statutory Basis Rubric \\
$\bullet$ Case Facts Rubric \\
$\bullet$ Reasoning Process Rubric \\
$\bullet$ Other Section Rubrics \\
\textbf{Response to be Evaluated:} The text requiring scoring, clearly labeled with the four sections: ``Conclusion,'' ``Statutory Basis,'' ``Case Facts,'' and ``Reasoning Process.'' \\
\midrule
\textbf{\textsc{Scoring Principles}} \\
1. \textbf{Strict Literal Matching:} Use the phrasing in the rubric as the standard; check whether the response contains identical or substantially identical expressions. \\
2. \textbf{Independent Evaluation:} Each scoring item is calculated independently without considering other items. \\
3. \textbf{No Inference or Supplementation:} Score based solely on what is explicitly mentioned in the response; do not perform reasoning or add information on the model's behalf. \\
4. \textbf{Clear Addition/Deduction:} Strictly follow the assigned values for adding or deducting points. \\
5. \textbf{Discretionary Partial Credit:} If the response does not fully cover a scoring point, award partial credit based on the key elements that were successfully addressed. \\
\midrule
\textbf{\textsc{Scoring Workflow}} \\
\textbf{Step 1: Segment the Response.} Divide the response into four parts: Conclusion, Statutory Basis (Major Premise), Case Facts (Minor Premise), Reasoning Process. \\
\textbf{Step 2: Individual Section Scoring.} For each part, use the corresponding rubric. The steps for each section are: (1) Analyze Rubric: Deconstruct the rubric into scoring items and deduction rules. (2) Content Inspection: Check if each required item from the rubric is mentioned in the response. \\
\textbf{Step 3: Structural Scoring (If applicable).} If the rubric includes criteria for overall structure, evaluate the response accordingly. \\
\textbf{Step 4: Summary and Feedback.} \\
\midrule
\textbf{\textsc{Output Format}} \\
Output a detailed evaluation for the following four parts, along with the total score and overall feedback. \\
\texttt{\{} \\
\hspace*{1em} \texttt{"score\_details": \{} \\
\hspace*{2em} \texttt{"Statutory Basis": \{} \\
\hspace*{3em} \texttt{"total\_points": 0, ``max\_points": 0,} \\
\hspace*{3em} \texttt{"breakdown": [ \{} \\
\hspace*{4em} \texttt{"rubric\_item": ``Description...", ``max\_points": 0, ``points\_awarded": 0,} \\
\hspace*{4em} \texttt{"mentions": ["Relevant snippets..."], ``rationale": ``Reasoning..."} \\
\hspace*{3em} \texttt{\} ]} \\
\hspace*{2em} \texttt{\}, ... [Structure repeated for Case Facts, Reasoning Process, Conclusion, Others]} \\
\hspace*{1em} \texttt{\},} \\
\hspace*{1em} \texttt{"total\_score": \{ ``total\_awarded": 0, ``total\_max": 0, ``percentage": ``0\%'' \},} \\
\hspace*{1em} \texttt{"overall\_feedback": \{ ``strengths": [], ``weaknesses": [], ``suggestions": [] \}} \\
\texttt{\}} \\
\midrule
\textbf{\textsc{User Input}} \\
The case analysis question is \texttt{<question><asking>} and its scoring rubric is \texttt{<rubrics>}. Please evaluate the following response: \texttt{<answers>}. \\
\bottomrule
\end{tabularx}
\caption{Judge prompt for scoring Task 2: Practical Case Analysis. }
\label{tab:score_prompt_task2}
\end{table*}
\clearpage
\twocolumn

\clearpage
\onecolumn
\begin{table*}[h]
\centering
\small
\renewcommand{\arraystretch}{1.25}
\begin{tabularx}{\linewidth}{>{\raggedright\arraybackslash}X}
\toprule
\textbf{Scoring Prompt 3: Legal Document Generation} \\
\midrule
\textbf{\textsc{Role}} \\
You are a legal practice expert specializing in the formal drafting standards of legal instruments. I will provide you with statements or other information from your client (the party involved) (\texttt{\$\{question\}}). Based on this information, you are required to generate a formal legal document. The document types and specific requirements are as follows: \\
\midrule
\textbf{\textsc{1. Document Types}} \\
Generate the appropriate legal document from the following options (ensure the content strictly adheres to the specific requirements of that type): \\
(1) Application (Bill of Complaint) \\
(2) Defendant (Statement of Defense) \\

\midrule
\textbf{\textsc{2. Document Requirements}} \\
(1) \textbf{Template Adherence:} Automatically draft the document according to the specific template provided for the given case. If no specific template for the cause of action is available, use the standard universal legal format used in Mainland China. \\
(2) \textbf{Length Constraint:} The legal document you draft must be between 2,500 and 3,000 characters. \\
(3) \textbf{Concise Output:} Output the document text only. Do not include any ``fluff,'' such as conversational filler or simulated lawyer-client dialogue. \\
(4) \textbf{Autonomous Selection:} You must independently determine which of the three document types is appropriate based on the provided prompt and context. \\
(5) \textbf{Critical Analysis:} Clients may have low legal literacy, and their statements may be inaccurate or contain biases. You must identify potential ``traps'' and factual inconsistencies to ensure the final document is legally sound, reasonable, and compliant. \\
(6) \textbf{Statutory Citations:} You must cite authentic and currently effective laws and regulations in the document. \\
\midrule
\textbf{\textsc{User Input}} \\
This is a scenario for legal document drafting: \texttt{<conversation>}. The evaluation criteria are \texttt{<rubrics>}. Please score the following legal document: \texttt{<answers>}. \\
\bottomrule
\end{tabularx}
\caption{Judge prompt for scoring Task 3: Legal Document Generation.}
\label{tab:score_prompt_task3}
\end{table*}

\clearpage
\twocolumn

\section{Evaluation Methodology Details}
\label{sec:app_evaluation}

\subsection{Judge Model Validation Results}
To identify the most accurate judge model, we calculated the alignment between scores from human experts and various LLMs to determine which model most closely approximates human judgment. The evaluation dataset consisted of the following:

\begin{enumerate}
    \item \textbf{Case Analysis:} A random 10\% sample of all items, paired with responses from a randomly selected model.
    \item \textbf{Document Drafting:} Ten cases each for Bill of Complaint and Statement of Defense, paired with responses from a randomly selected model.
    \item \textbf{Legal Consultation:} Responses to 20 test items generated by four specific models (Claude 3.5 Sonnet, GPT-5-0807-Global, Qwen-Max, and Gemini 3 Pro Preview).
\end{enumerate}

These responses were distributed to both human experts and LLMs for scoring. To ensure reliability, each response was evaluated by at least three human experts. We compared individual expert scores against the group average; the Pearson correlation coefficients between the three experts' scores and the human average ranged from $0.8391$ to $0.8841$, with a Mean Absolute Error (MAE) between $5.1937$ and $7.9817$. Table \ref{tab:judge_alignment} presents the alignment results between candidate Judge models and human experts. 

Following current research trends that employ high-performance models as evaluators, we tested Gemini 3 Pro Preview, GPT-5.1-1113-Global, and Qwen3-Max as potential judge models. Based on their comprehensive performance across different tasks, we selected \textbf{Gemini 3 Pro Preview} as our final judge model. Although it did not achieve the highest alignment in the Document Drafting task, it demonstrated superior performance in the remaining two categories.

\subsection{Scoring Mechanics}
To prevent scoring discrepancies, we utilize a scoring rate (Earned Points / Total Points). We established eight comprehensive metrics. Since LLMs are prone to calculation errors, we employ regular expressions to extract raw scores from the JSON or list outputs generated by the Judge model (as defined in the scoring prompts) and perform the final summation programmatically.

\section{Detailed Error Analysis}
\label{sec:app_error_analysis}

\subsection{Error Analysis of Public Legal Consultation}
For Task 1, we analyzed 183 low-scoring responses and identified four common categories of errors along with eight specific issues. The common problems observed in current LLMs during public legal consultation include:
\begin{itemize}
    \item \textbf{Insufficient grasp of facts and causal relationships.} Models often struggle to accurately link reported events to their legal consequences.
    \item \textbf{Shallow understanding of evidence.} Models frequently fail to build a closed logical loop of questioning around evidence, missing the necessity of corroborating key facts.
    \item \textbf{Weak identification of specific evidence types.} There is a general lack of sensitivity to the specific forms of evidence required for different legal claims.
    \item \textbf{Defects in role adherence.} Some models exhibit limitations in their ability to discern and analyze the overall case context due to persona setting issues.
\end{itemize}

Beyond these commonalities, distinct model-specific weaknesses were observed. For instance, \textbf{GPT-5.2} tends to deviate from key facts during follow-up questioning, resulting in divergent lines of inquiry. \textbf{Gemini-3.0} struggles with cases involving complex evidence, often failing to understand probative requirements and asking questions without sufficient legal basis.

Overall, the performance of current LLMs in simulating real-world legal consultation remains insufficient. Whether in fact organization, causal inference, or evidence analysis, models generally exhibit shallow understanding and fragmented reasoning. This is particularly evident in complex cases with multiple causes or weak evidence chains, where misjudgments and omissions of key points are frequent.

\subsection{Error Analysis of Practical Case Analysis}
In Task 2, we examined 200 low-scoring cases and summarized the errors into four dimensions covering 16 specific issues. The common issues across models are:

\begin{itemize}
    \item \textbf{Conclusion:} Fundamental errors in legal characterization. 
    \item \textbf{Case Facts:} Omission of key facts that affect qualitative analysis, as well as loss of important details such as time, amount, and subject identity. 
    \item \textbf{Reasoning Process:} Failure to follow the required legal reasoning sequence (e.g., syllogism). 
    \item \textbf{Statutory Basis:} Absence of core legal provisions.
\end{itemize}

Model-specific analysis reveals that \textbf{GPT-5.2} has significant deficiencies in legal retrieval and application, frequently missing important legal bases or citing incorrect versions. \textbf{Gemini-3.0} shows defects in the integrity of reasoning chains and logical rigor, alongside serious issues with missing core statutes.

In summary, current LLMs have not yet reached a professional level in practical case analysis. While they can generate structured text, they exhibit obvious shortcomings in verifiable analysis: often omitting key facts, reasoning with improper order or broken chains, and misapplying laws (e.g., mixing new and old laws, or failing to distinguish between general and special provisions).

\vspace{1em}
\noindent\fbox{
\begin{minipage}{0.95\linewidth}
\textbf{Case Study: Dispute over Cryptocurrency Mining Machine Sales}

\textbf{Case Summary:} Party A purchased ``Innosilicon AL0 Servers'' (mining machines) from Party B for 50,000 RMB (48,000 RMB for goods + 2,000 RMB shipping). Party B labeled them ``no returns or exchanges.'' Party A refused the initial delivery. A second notarized delivery confirmed the goods did not match the agreed model, and Party A refused again. Party B neither refunded nor replaced the goods. Both parties confirmed the subject matter was mining machines. Party B knew that ``the state does not allow sales,'' and Party A was a ``miner'' aware of the usage.

\textbf{Question:} Answer with logic [Conclusion] + [Case Facts] + [Reasoning] + [Statutes]: Should the court order Party B to return 50,000 RMB and corresponding interest?

\textbf{Analysis of Model Performance:}
The model exhibited significant failures in sequential reasoning:
\begin{itemize}
    \item \textbf{Reasoning Order Violation:} The model skipped the mandatory first step of analyzing ``contract validity'' and directly proceeded to ``breach of contract,'' assuming the contract was valid. It failed to follow the ``Step 1 (Validity) $\to$ Step 2 (Return) $\to$ Step 3 (Interest)'' sequence.
    \item \textbf{Missing Core Premise:} It failed to identify the illegality of trading mining machines, erroneously citing ``no explicit legal prohibition'' to deem the contract valid. This ignores regulatory restrictions and public order requirements.
    \item \textbf{Broken Logical Chain:}
    \begin{itemize}
        \item In Step 2, instead of applying ``void contract restitution'' (Civil Code Art. 157), it applied ``contract rescission for breach.'' It also failed to separate shipping costs from the goods price.
        \item In Step 3, it supported the interest claim without analyzing the mutual fault regarding the illegal transaction, ignoring that both parties share responsibility for the void contract.
    \end{itemize}
\end{itemize}
\textbf{Diagnosis:} The error stems from an inability to prioritize the ``validity judgment'' as a prerequisite. By skipping this foundational step, the subsequent reasoning on liability and remedies lost its legal basis.
\end{minipage}
}

\subsection{Error Analysis of Legal Document Generation}
For Task 3, an analysis of 183 low-scoring responses revealed two common issues:
\begin{itemize}
    \item \textbf{Inaccurate Legal Application:} Insufficient grasp of the timeliness, relevance, and citation principles of laws.
    \item \textbf{Weak Jurisdiction Review:} Insufficient sensitivity and capability in reviewing jurisdictional rules.
\end{itemize}
These problems are prevalent across all models. Both international state-of-the-art models (e.g., GPT-5.2, Claude-4.5) and other strong models (e.g., DeepSeek, Qwen) struggle to perfectly resolve issues related to legal "hallucinations" and procedural logic.

Regarding individual differences, \textbf{GPT-5.2}, \textbf{Claude-4.5}, and \textbf{Gemini-3.0} make fewer errors in basic legal logic such as "proper party" determination and "fact summarization," indicating stronger instruction following and stability. Conversely, models like \textbf{DeepSeek}, \textbf{Kimi}, and \textbf{Qwen3-Max} are more prone to hallucinations and blindly following client instructions even when they are legally unsound.

Overall, while current models are mature in text organization, they lack substantive quality. They often fail to judge the suitability of subjects (blindly following wrong instructions), lack sensitivity to the statutory scope of claims (accepting unreasonable claims or omitting lawful ones), and violate the rigor required for legal documents.

\clearpage

\stopcontents[appendix]

\end{document}